\documentclass{article}

 \usepackage[eandd, final]{neurips_2026}
\usepackage[utf8]{inputenc} % allow utf-8 input
\usepackage[T1]{fontenc}    % use 8-bit T1 fonts
\usepackage[colorlinks=true, citecolor=orange, linkcolor=magenta]{hyperref}       % hyperlinks
\usepackage{url}            % simple URL typesetting
\usepackage{arydshln} 
\usepackage{booktabs}       % professional-quality tables
\usepackage{amsfonts}       % blackboard math symbols
\usepackage{nicefrac}       % compact symbols for 1/2, etc.
\usepackage{microtype}      % microtypography
\usepackage[dvipsnames, table]{xcolor}         % colors
\usepackage[font=footnotesize]{caption}
\usepackage{enumitem}
\definecolor{airforceblue}{rgb}{0.36, 0.54, 0.66}
\definecolor{aquamarine}{rgb}{0.5, 1.0, 0.83}
\definecolor{babyblue}{rgb}{0.54, 0.81, 0.94}
\definecolor{babyblueeyes}{rgb}{0.63, 0.79, 0.95}
\definecolor{blue(munsell)}{rgb}{0.0, 0.5, 0.69}
\definecolor{blue(ncs)}{rgb}{0.0, 0.53, 0.74}
\definecolor{ceruleanblue}{rgb}{0.16, 0.32, 0.75}
\definecolor{cobalt}{rgb}{0.0, 0.28, 0.67}
\definecolor{darkcerulean}{rgb}{0.03, 0.27, 0.49}
\definecolor{darkpowderblue}{rgb}{0.0, 0.2, 0.6}
\definecolor{denim}{rgb}{0.08, 0.38, 0.74}
\definecolor{egyptianblue}{rgb}{0.06, 0.2, 0.65}

\usepackage{algorithm}
\usepackage{algorithmic}

\usepackage{amsmath}
\usepackage{amssymb}
\usepackage{mathtools}
\usepackage{amsthm}
\usepackage{multirow}
\usepackage{subcaption}
\usepackage{array}
\usepackage{makecell}

\usepackage{wrapfig}
\usepackage{booktabs}
\usepackage{tabularx}
\usepackage[flushleft]{threeparttable}
\usepackage{colortbl}
 \usepackage{transparent}
\usepackage{makecell}  % Optional: for better cell formatting
\newcommand\Tstrut{\rule{0pt}{2.6ex}}         % = `top' strut
\usepackage[capitalize,noabbrev]{cleveref}
\newcommand{\mstd}[2]{#1 {\scriptsize $\pm$#2 }}
\newcommand{\mstdgray}[2]{\textcolor{black!60}{#1} {\scriptsize $\pm$\textcolor{black!60}{#2} }}
\newcommand{\conf}[1]{\textcolor{gray}{\scriptsize #1}}

\theoremstyle{plain}

\theoremstyle{definition}

\theoremstyle{remark}

\usepackage{mathtools}

\title{Rethinking Dataset Distillation for Classification: \\ Do Distilled Sets Outperform Coresets?}

\author{Trisha Mittal\textsuperscript{1*}, Akshay Mehra\textsuperscript{1*}, and Joshua Kimball\textsuperscript{1}\\
{\small \textsuperscript{*}Equal Contribution \textsuperscript{1}Dolby Laboratory}\\ 
{\tt\small\{trisha.mittal, akshay.mehra, 
joshua.kimball\}@dolby.com}\\
}

\begin{document}

\maketitle

\begin{abstract}
Dataset distillation (DD) has emerged as a prominent approach in data centric machine learning, aiming to \emph{synthesize} compact training sets for efficient training by compressing the information in large datasets into a small number of synthetic samples. 
However, DD methods are often evaluated under inconsistent evaluation protocols, ranging from standard ERM to single/multi‑teacher supervision, making it difficult to isolate the effectiveness of distilled data from evaluation. 
Moreover, many prior methods claim that DD outperforms data pruning approaches such as coreset selection (CS), based on the assumption that restricting condensed datasets to subsets of real samples fundamentally limits their expressiveness. 
In this work, we critically evaluate DD methods through large-scale experiments using standardized datasets and evaluation protocols to assess their intrinsic effectiveness.
We benchmark \emph{seven} state-of-the-art (SOTA) DD methods on ImageNet-1K, ImageNet-100, and ImageNette, using \textit{three} widely adopted training protocols against \emph{four} CS strategies. 
Our results show that while some DD methods fail to outperform even simple random subsets, the SOTA DD approaches are comparable to or worse than coresets on large‑scale datasets and incur a substantially higher cost for construction. 
Beyond accuracy, we also evaluate the representativeness, diversity, and quality of condensed sets, and find that coresets consistently achieve better coverage of the original data distribution. 
These findings highlight the limited practical advantages of current DD methods and show that coresets remain competitive and are often a more computationally efficient alternative for data-centric learning.
\end{abstract}
\section{Introduction}
Recent progress in machine learning has been driven primarily by scaling data, models, and computation, leading to the success of large language and multimodal models in a wide range of domains~\cite{touvron2023llama, achiam2023gpt}. 
However, this approach comes with significant computational, storage, and environmental costs that challenge its long‑term sustainability. 
In response, there has been a growing shift toward data-centric approaches that prioritize the quality, structure, and utility of training data over sheer scale ~\cite{sorscher2022beyond, li2024datacomp}.
% Recent studies~\cite{sorscher2022beyond, li2024datacomp} show that well‑chosen, informative, and diverse subsets can match or exceed the performance of full‑dataset training at substantially lower computational cost.

% In recent years, the machine learning community has primarily advanced model performance by scaling data and computation, culminating in the success of large language and multimodal models (LLMs and MLLMs) across diverse domains. 
% Although effective, this paradigm incurs prohibitive computational, storage, and environmental costs, raising questions about its long-term sustainability. 
% Consequently, there has been a growing shift towards data-centric approaches that emphasize the quality, structure, and utility of training data rather than sheer scale. 
% Recent work demonstrates that carefully identifying informative and diverse subsets can match or even surpass the performance of models trained on full datasets, while substantially reducing training cost. 
% These advances suggest that the more judicious use of data through improved curation, selection, and representation is a promising and sustainable alternative for further progress in machine learning.

Within data‑centric machine learning dataset distillation (DD) \cite{liu2025evolution,shang2025dataset} has emerged as an effective approach, demonstrating that compact synthetic training sets can recover strong performance despite extreme compression. 
However, evaluations of DD methods often depend on specific training pipelines with some works using standard ERM while others adopt student/teacher–based training making it difficult to disentangle whether reported performance gains arise from improved data representations or from evaluation‑specific design choices. 
% As a result, it remains unclear whether the reported gains reflect fundamental advantages of distillation or are instead artifacts of evaluation design and benchmarking choices.
% Since DD methods optimize the distilled sets as continuous parameters, DD methods are regarded as more expressivethan approaches that merely select subsets of the original data. 

Alongside this line of work, coreset selection (CS) \cite{mirzasoleiman2020coresets,guo2022deepcore} has also been studied extensively, showing that carefully chosen real examples can provide effective condensed training sets. 
However, in many recent DD works~\cite{zhao2025taming,chan2025mgd3,roy2026manifoldgd}, CS methods are frequently portrayed as inherently limited, considered ineffective at high compression, or are omitted from empirical comparisons. 
Such claims of DD’s superiority over CS are often made without systematic evaluation against strong CS baselines, particularly on large‑scale datasets and across multiple evaluation protocols. 
Another limitation of DD methods is their reliance on pretrained generative models trained on the full dataset, which can make them difficult or infeasible to extend to datasets where such models are unavailable. 
In contrast, CS methods do not depend on pretrained generative models, reinforcing their role as a necessary point of comparison when evaluating data‑efficient learning approaches.
These concerns echo a broader pattern documented in the reproducibility literature, where reported gains often diminish once evaluation protocols are standardized, undocumented design choices are controlled for, or simple but strong baselines are restored~\cite{gundersen2018state, gundersen2022sources, raff2019step, belz2021systematic}.

\begin{figure}[t]
    \centering
    \begin{subfigure}[t]{0.48\linewidth}
        \centering
        \includegraphics[width=\linewidth]{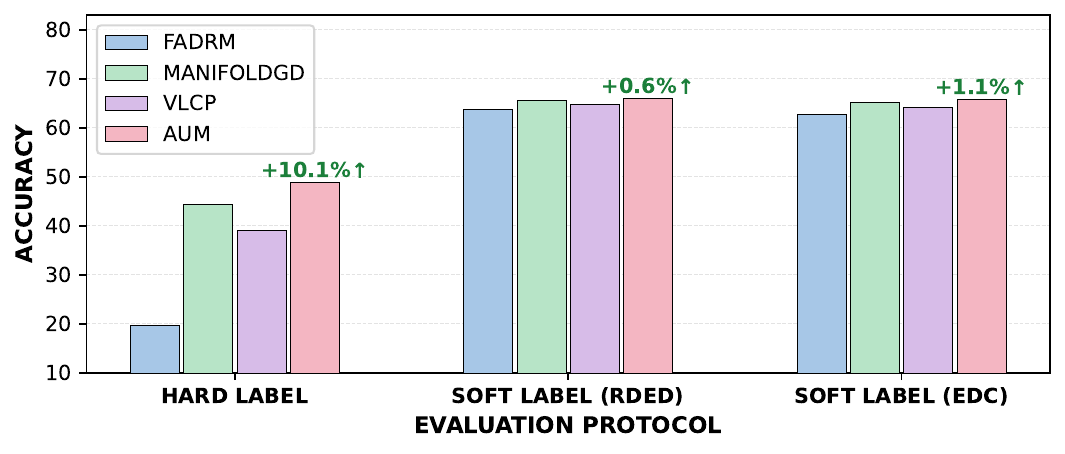}
        \caption{}
        \label{fig:teaser_subfig1}
    \end{subfigure}
    \hspace{-20pt}
    \hfill
    \begin{subfigure}[t]{0.28\linewidth}
        \centering
        \includegraphics[width=\linewidth]{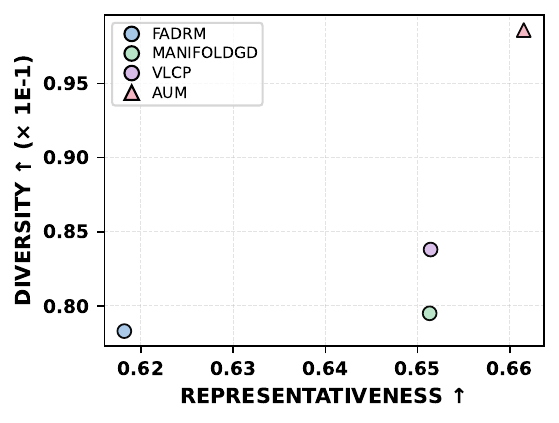}
        \caption{}
        \label{fig:teaser_subfig2}
    \end{subfigure}
    \hspace{-20pt}
    \hfill
    % \begin{subfigure}[t]{0.22\linewidth}
    %     \centering
    %     \includegraphics[width=\linewidth]{images/imagenet1k_fid_barplot_ipc50.pdf}
    %     \caption{}
    %     \label{fig:subfig3}
    % \end{subfigure}
        \begin{subfigure}[t]{0.21\linewidth}
        \centering
        \includegraphics[width=\linewidth]{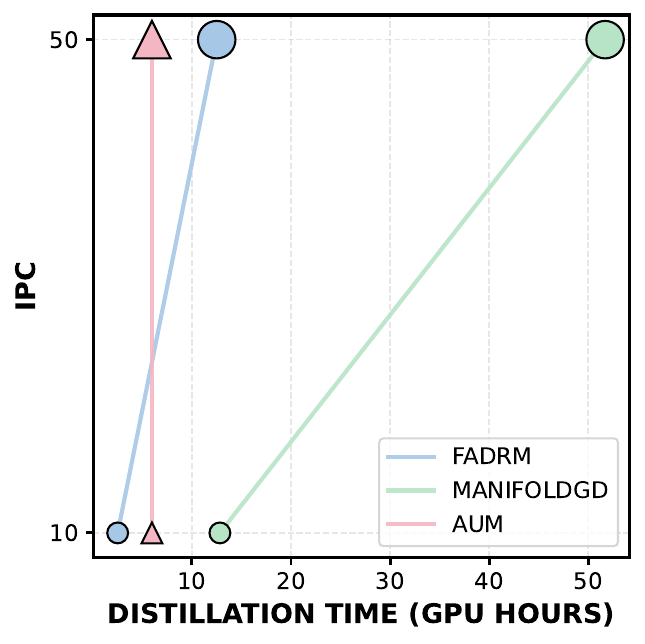}
        \caption{}
        \label{fig:teaser_subfig3}
    \end{subfigure}
    \vspace{-0.25cm}
    \caption{
    % \small{\textbf{\AM{verify percent improvement numbers, mention 50IPC in the table, which methods are DD and CS} Coreset Selection vs. Dataset Distillation Baselines.} Our results show that coreset selection baselines outperform a wide range of dataset distillation methods. (a) Top-1 accuracy comparison between coreset selection and dataset distillation on ImageNet-1K. (b) Coresets exhibit greater diversity and better representativeness than distilled datasets. (c) The quality of distilled datasets, measured using Fréchet Inception Distance (FID), is inferior to that of coresets. \AM{mention why we don't have VLCP}}
    \textbf{Coresets consistently outperform distilled sets  in accuracy, data coverage, and construction efficiency.} We report results on ImageNet-1K using condensed sets consisting of 50 images per class.
    (a) Relative accuracy gains of a ResNet-50 trained on coreset identified using AUM \cite{pleiss2020identifying} in comparison to three SOTA DD methods (FADRM+ \cite{cui2026fadrm}, ManifoldGD \cite{roy2026manifoldgd}, VLCP \cite{zou2025vlcp}) for three evaluation protocols.
    (b) Representativeness–diversity comparison showing broader coverage of the data distribution provided by coresets~(top-right corner).
    (c) Wall‑clock construction cost versus IPC with constant scaling for CS methods versus linear scaling for  DD, highlighting the efficiency advantage of CS.
    }
    \label{fig:teaser_fig}
    \vspace{-15pt}
\end{figure}

In the same spirit, and motivated by concerns about both the sensitivity of DD performance to evaluation pipelines and the absence of systematic comparisons with CS baselines, we present a systematic study of CS and DD under standardized datasets and evaluation protocols. 
We pose two core questions: \emph{does the success of DD persist when disentangled from evaluation pipelines, and do distilled datasets provide inherent advantages over coresets under controlled evaluation?}
To answer these questions, we perform a systematic evaluation of \emph{seven} SOTA DD methods, covering three generative training‑based approaches~\cite{gu2024efficient, zou2025vlcp, zhao2025taming}, three generative training‑free approaches~\cite{hinton2015distilling, chan2025mgd3, roy2026manifoldgd}, and one non‑generative DD method~\cite{cui2026fadrm} (see Table~\ref{tab:related_methods} for a summary).
We compare these DD methods against \emph{four} widely used CS methods, including random sampling, clustering, accumulated margin (AUM) based selection~\cite{pleiss2020identifying}, and concept‑guided CS~\cite{mehra2025coreset}. 
All evaluations are conducted on ImageNet‑1K, ImageNet‑100, and ImageNette using three standard evaluation pipelines (hard label: where models are directly trained on condensed sets and soft label where one (RDED \cite{sun2024diversityrealismdistilleddataset}) or more (EDC \cite{shao2024elucidating}) teacher models are used for knowledge distillation) commonly adopted in DD works. %, allowing a controlled and protocol‑aware comparison between CS and DD.

% Our findings in Fig.~\ref{fig:teaser_fig} show that SOTA DD approaches perform on par with or slightly worse than coresets. 
% As shown in Fig.~\ref{fig:subfig1}, CS via AUM consistently achieves the highest accuracy across all three evaluation protocols.
% The advantage is most pronounced under the hard‑label protocol indicating that distilled datasets do not provide intrinsic advantage over coresets. 
% While supervision from teacher models under the soft‑label protocol narrows the gap across CS and DD methods, the best DD method (e.g., ManifoldGD which requires a generative model trained on the full dataset) only matches rather than exceed the performance of CS, despite using large computational resources. 
% Moreover, our results in Sec.~\ref{sec:experiments} show that no single DD method consistently performs the best across datasets and evaluation protocols, underscoring sensitivity of DD methods to evaluation pipeline. 

Our findings show that state-of-the-art DD methods perform on par with, or worse than, coreset-based approaches (Fig.~\ref{fig:teaser_fig}(a)), especially on large-scale datasets, despite using large computational resources (e.g., for training generative models before distillation). 
Moreover, CS methods consistently achieve a more favorable representation--diversity trade-off compared to DD methods (Fig.~\ref{fig:teaser_fig}(b)) showing better distribution coverage. We also observe a substantial gap in construction cost between these two families of methods, with DD approaches being significantly more expensive (Fig.~\ref{fig:teaser_fig}(c)). 
Moreover, our results in Sec.~\ref{sec:experiments} show that no single DD method consistently performs the best across datasets and evaluation protocols, underscoring sensitivity of DD methods to evaluation pipeline.
Overall these results indicate that success of DD needs to be reevaluated especially in comparison to CS which provides a more efficient and scalable solution in practice. 
The key takeaways from this work are: 
% \AM{Read the eval track call and make these aligned to that.}

\begin{itemize}[leftmargin=0.5cm, noitemsep]
    \item Through our extensive empirical analysis, we find that 1) performance of DD methods are entangled with the evaluation pipeline and 2) DD methods do not exhibit an intrinsic advantage over CS when evaluated under standardized datasets and training pipelines. 
    
    \item In our analysis of condensed sets with respect to representativeness, diversity, and computational cost, we find that coresets achieve higher accuracy and broader distributional coverage while incurring substantially lower construction cost.
    
    \item Our results establish CS as a necessary baseline for future DD methods and highlight the importance of adopting standardized evaluation protocols, including the use of hard labels alongside soft-label approaches based on single/multiple teachers for benchmarking future DD methods.

    % \item Our results establish CS as a necessary baseline for future DD methods and highlight the importance of adopting standardized evaluation protocols, including the use of hard labels alongside soft-label approaches based on single/multiple teachers for benchmarking future DD methods, in line with broader calls for rigorous and reproducible evaluation in ML~\cite{gundersen2022sources, raff2019step}.
    % \item Our extensive empirical analysis with SOTA DD methods, demonstrates that existing DD methods do not consistently outperform CS across datasets or evaluation protocols, establishing CS as a necessary baseline for evaluating the success of future DD methods.

\end{itemize}

\section{Background and Related Work}

\begin{table*}[t]
\centering

\caption{\textbf{Taxonomy for CS and DD Methods.} We present a taxonomy to summarize the various coreset selection and dataset distillation methods in the last few years including the method requirements and assumptions. 
% \AM{Can we move CS methods to a separate row, it's confusing right now.} 
}
\label{tab:related_methods}

\begin{threeparttable}
{\renewcommand{\arraystretch}{1.25}
\resizebox{0.99\textwidth}{!}{
\begin{tabular}{
m{.05\linewidth} 
p{.16\linewidth} 
@{\hspace{0.5cm}} 
m{.34\linewidth} 
>{\centering\arraybackslash}m{.12\linewidth} 
>{\centering\arraybackslash}m{.12\linewidth} 
>{\centering\arraybackslash}m{.12\linewidth} 
>{\centering\arraybackslash}m{.12\linewidth} 
>{\centering\arraybackslash}m{.12\linewidth}
>{\centering\arraybackslash}m{.12\linewidth}
}

\toprule[1.25pt]

\multicolumn{2}{c}{\raisebox{-0.2cm}{\textbf{Method type}}}  &
\raisebox{-0.2cm}{\textbf{Method name (venue)}} &
\textbf{Needs full training data?} &
\textbf{Requires generative model trained on full data?} &
\textbf{Purpose of generative model?} &
\textbf{Construction cost scales with IPC} &
\textbf{Requires a classifier trained on full data?} &
\textbf{Requirement for extending to new data?} \\

\midrule

% \multirow{8}{*}{\raisebox{-0.8cm}{\rotatebox[origin=c]{90}{ \textbf{Generative}}}}
\multirow{10}{*}{\rotatebox[origin=c]{90}{\makecell[c]{\textsc{\textbf{DD}}\\ \textsc{Methods}}}}&

\multirow{1}{*}{\raisebox{-0.1cm}{ \makecell[c]{\textbf{Generative}\\ \textbf{Training-based}}}} &
\colorbox{Cerulean!30}{Minimax} (\conf{CVPR'24})~\cite{gu2024efficient},
\colorbox{Cerulean!30}{VLCP} (\conf{ICCV'25})~\cite{zou2025vlcp}, 
\colorbox{Cerulean!30}{D3HR} (\conf{ICML'25}),
D4M (\conf{CVPR'24})~\cite{Su_2024_CVPR},
Li et al. (\conf{ICCV'25 Wkshp})~\cite{li2025diff}
&
$\checkmark$ & \makecell[c]{$\checkmark$\\{ (often} \\ {DiT-XL/2)}}   & \makecell[c]{{ finetuning}} &   \makecell[c]{{ $\checkmark$}}& $\times$ &  \makecell[c]{{ Train} \\ { generative }\\{model}} \\

\cmidrule(ll){2-9}

&
\multirow{2}{*}{\raisebox{-0.8cm}{\makecell[c]{\textbf{Generative}\\ \textbf{Training-Free}}}} &
\colorbox{Cerulean!30}{DiT-Distillation} \cite{jin2024fast}
&
$\checkmark$ &\makecell[c]{$\checkmark$\\{ (often} \\ {DiT-XL/2)}}  & \makecell[c]{{ sampling}} &  \makecell[c]{{ $\checkmark$}}& $\times$ &  \makecell[c]{{ Train} \\ { generative }\\{model}}\\
\cdashline{3-9}
&&
\Tstrut\Tstrut\colorbox{Cerulean!30}{MGD$^3$} (\conf{ICML'25})~\cite{chan2025mgd3},
\colorbox{Cerulean!30}{ManifoldGD} (\conf{CVPR'26})~\cite{roy2026manifoldgd}
&
$\checkmark$ & \makecell[c]{$\checkmark$\\{ (often} \\ {DiT-XL/2)}} & \makecell[c]{{ guidance} \\ { during}\\ {sampling}}  &   \makecell[c]{{ $\checkmark$}} & $\times$ & \makecell[c]{{ Train} \\ { generative }\\{model}}\\

% \midrule
\cmidrule(ll){2-9}

% \multirow{1}{*}{\raisebox{15cm}{\rotatebox[origin=c]{90}{ \textbf{{Non-Generative}}}}} 
&
\makecell[c]{\textbf{Decoupled}\\\textbf{(SRe2L-based)}} &
\colorbox{Cerulean!30}{FADRM+} (\conf{NeurIPS'25})~\cite{cui2026fadrm},
EDC (\conf{NeurIPS'24})~\cite{shao2024elucidating},
RDED (\conf{CVPR'24})~\cite{sun2024diversityrealismdistilleddataset},
G-VBSM (\conf{CVPR'24})~\cite{shao2024generalizedlargescaledatacondensation}, 
SRe2L (\conf{NeurIPS'23})\cite{yin2023squeeze}
&
$\checkmark$ & $\times$ & -- & \makecell[c]{{ $\checkmark$}} & \makecell[c]{$\checkmark$} & \makecell[c]{{ Train}\\ { classifier}}\\

% \cmidrule(lr){2-8}
\midrule
\multirow{5}{*}{\rotatebox[origin=c]{90}{\makecell[c]{\textsc{\textbf{CS}}\\ 
\textsc{Methods}}}}

& \multirow{1}{*}{\makecell[c]{\textbf{Geometry-based}}} & {Clustering} & $\checkmark$ & $\times$ & -- & \makecell[c]{{ $\times$}} & \makecell[c]{{ $\times$}} & \makecell[c]{{Recompute} \\ {features and} \\ {clusters}}\rule[-2.9ex]{0pt}{0pt}\\

\cmidrule(ll){2-9}

& \multirow{3}{*}{\makecell[c]{\textbf{Score-based}}} & {AUM}~(\conf{NeurIPS'20})~\cite{pleiss2020identifying} & $\checkmark$ & $\times$ & -- & \makecell[c]{{ $\times$}} & \makecell[c]{$\checkmark$} & \makecell[c]{{ Train} \\ { classifier}}\rule[-2.9ex]{0pt}{0pt}\\
% \cdashline{3-9}

&&
{Concepts}~\cite{mehra2025coreset}
&
$\checkmark$ & $\times$ & -- &\makecell[c]{{ $\times$}} & \makecell[c]{$\checkmark$\\{ (small)}} & \makecell[c]{{ Train} \\{ classifier} \\ {(small)}}\\

\bottomrule[1.25pt]
\end{tabular}
}
{ \begin{tablenotes}[flushleft]
\item[\colorbox{Cerulean!30}{}] {\footnotesize Methods with open-sourced code available and/or released distilled datasets for ImageNet-1K.}
\end{tablenotes}}
}
\end{threeparttable}
% \vspace{-15pt}
\end{table*}
% Within data-centric machine learning, dataset distillation and coreset selection have emerged as two prominent paradigms for improving data efficiency, aiming to reduce the size of the data set while preserving representative information and downstream performance. We formally define their objectives and give an overview of previous work in these areas below (see Fig \ref{fig:overview-fig}). 
While DD and CS both compress large datasets while preserving performance, they differ in their assumptions, construction cost, and reliance on auxiliary models (Table~\ref{tab:related_methods}) which we overview here.

{\bf Preliminaries.} 
Consider a classification task with data distribution $\mathcal{P}$. 
Let the original dataset denoted by $\mathcal{D} = \{(x_i, y_i)\}^N_{i=1}$ be sampled i.i.d. from the distribution $\mathcal{P}$, where $x_i$ denotes the data and $y_i \in \mathcal{Y}$ denotes the label from a set of $c$ classes. Let $f_{\theta_\mathcal{D}}$ denote a model trained on $\mathcal{D}$, obtained through empirical risk minimization (ERM) over a loss function $\ell$, i.e., $\theta_\mathcal{D} = \arg \min_{\theta} \mathbb{E}_{(x,y) \in \mathcal{D}}[\ell(f_{\theta}(x), y)]$.

{\bf Dataset distillation (DD)}
constructs a {\bf ``synthetic''} dataset denoted by $\mathcal{S} = \{(\Tilde{x}_j, \Tilde{y}_j)\}^{M_S}_{j=1}$, with $M_S \ll N$ such that models trained on $\mathcal{D}$ and $\mathcal{S}$ exhibit the same generalization behavior. 
Oftentimes the distilled dataset is constructed such that images per class (IPC) are constrained to a small number such as 10 or 50.
Mathematically, the problem of dataset distillation is 
\begin{equation}
\label{eq:distillation}
% \small
\min_{\mathcal{S},|\mathcal{S}|=M_S} \big| \mathbb{E}_{(x,y) \sim \mathcal{P}}[\ell(f_{\theta_\mathcal{D}}(x),y)] - \mathbb{E}_{(x,y) \sim \mathcal{P}}[\ell(f_{\theta_\mathcal{S}}(x),y)] \big|,
\end{equation}
where $f_{\theta_\mathcal{S}}$ denotes a model trained on the $\mathcal{S}$ via ERM, i.e., $\theta_\mathcal{S} = \arg \min_{\theta} \mathbb{E}_{(\Tilde{x}, \Tilde{y}) \in \mathcal{S}}[\ell(f_{\theta}(\tilde{x}), \tilde{y})]$.
% Recent advances in dataset distillation have led to a diverse set of approaches, particularly in the large-scale regime. 
% As summarized in Table~\ref{tab:related_methods}, existing DD methods can be broadly categorized into {\em generative} and {\em non-generative} approaches. %, with further distinctions based on whether they require explicit training or operate in a training-free manner.
% As summarized in Table~\ref{tab:related_methods}, existing DD methods differ substantially in their assumptions and computational requirements. 
% A central distinction is whether methods rely on a generative model, and if so, how this model is used during dataset construction.

As summarized in Table~\ref{tab:related_methods}, a central distinction between DD methods is whether these methods rely on a generative model, and if so, how this model is used for construction of distilled sets.
Generative DD approaches leverage diffusion models or GANs to synthesize condensed datasets, while non‑generative approaches directly optimize or reconstruct samples (see App.~\ref{app:additional_rw}).

Within \textit{generative approaches}, a key distinction is between {\em training-based} and {\em training-free} approaches. 
Training-based methods such as D$^4$~\cite{Su_2024_CVPR} and Minimax~\cite{gu2024efficient}, fine-tune diffusion model to obtain distilled datasets. % that improve representativeness and downstream performance. 
% explicitly train or fine-tune the generative model to produce informative synthetic data. 
% Representative works include  fine-tune diffusion model to obtain distilled datasets that improve representativeness and downstream performance.
More recent methods such as VLCP~\cite{zou2025vlcp}, D3HR~\cite{zhao2025taming}, and task-specific generative distillation approaches~\cite{li2025diff} incorporate semantic priors, difficulty-aware sampling, or vision-language supervision to enhance the quality of distilled data. 
While these approaches achieve strong performance on large-scale datasets, they often require substantial training/fine-tuning cost. % and careful fine-tuning.
In contrast, {\em training-free methods} decouple dataset synthesis from expensive optimization by using pretrained generative models. 
Approaches such as DiT Distillation \cite{jin2024fast}, MGD$^3$~\cite{chan2025mgd3}, and ManifoldGD~\cite{roy2026manifoldgd} guide the sample during the reverse diffusion step to generate a more diverse set of distilled samples.
These methods significantly reduce computational overhead while maintaining competitive performance, making them particularly attractive for large-scale settings. 
However, their effectiveness depends on the availability, quality and coverage of the generative model for a domain.

Some \textit{non-generative approaches} formulate %construct distilled datasets directly without relying on generative models. 
% A prominent class of \textit{Non-generative approaches} early methods formulates
distillation as a bi-level problem, where synthetic data is learned to match gradients or training trajectories of real data~\cite{zhao2021dataset,cazenavette2022dataset}. 
While effective at small scales, these approaches do not scale well due to high computational and memory requirements.
% To address this limitation, {\em decoupled methods} have emerged as a dominant paradigm for large-scale distillation. 
Other approaches %in this paradigm %These approaches 
separate dataset construction from model training, leveraging pretrained classifiers to reconstruct informative samples or align intermediate statistics. 
Representative methods include SRe2L \cite{yin2023squeeze} and its extensions such as  RDED~\cite{sun2024diversityrealismdistilleddataset}, G-VBSM~\cite{shao2024generalizedlargescaledatacondensation},  EDC~\cite{shao2024elucidating}, and FADRM+~\cite{cui2026fadrm}. 
While these methods %By avoiding costly bi-level optimization, these methods 
offer improved scalability and efficiency, %although 
they suffer from limited diversity or information loss due to the separation between model knowledge and data synthesis.

{\bf Coreset selection (CS)} aims to identify a {\bf ``subset''} of the original dataset denoted by $\mathcal{C} = \{(x_k, y_k)\}^{M_C}_{k=1}$, with $M_C \ll N$ such that models trained on $\mathcal{D}$ and $\mathcal{C}$ exhibit the same generalization behavior. 
Formally, the problem of coreset selection is 
\begin{equation}
\label{eq:coreset}
% \small
\min_{\mathcal{C}:\mathcal{C}\subset\mathcal{D},\left\lvert\mathcal{C}\right\rvert=M_C} \left\lvert \mathbb{E}_{(x,y) \sim \mathcal{P}}[\ell(f_{\theta_\mathcal{D}}(x),y)] - \mathbb{E}_{(x,y) \sim \mathcal{P}}[\ell(f_{ \theta_\mathcal{C}}(x),y)] \right\rvert,
\end{equation}

where $f_{\theta_\mathcal{C}}$ denotes a model trained on the $\mathcal{C}$ via ERM i.e., $\theta_\mathcal{C} = \arg \min_{\theta} \mathbb{E}_{(x, y) \in \mathcal{C}}[\ell(f_{\theta}(x), y)]$.
A prominent class of CS methods relies on {\em training dynamics} of a classification model to score samples. 
These include approaches based on forgetting events~\cite{toneva2018empirical}, margin-based scores such as AUM~\cite{pleiss2020identifying}, and entropy-based uncertainty sampling~\cite{coleman2019selection}. 
While effective, these methods require training a (proxy or downstream) model on the full dataset at least once to compute scores, making them computationally expensive for large-scale datasets.  
In contrast, recent work~\cite{mehra2025coreset} estimates sample importance using concept bottlenecks, providing a model-agnostic, interpretable and efficient alternative to training dynamics-based methods. 
% In this method, concepts extracted via LLMs are aligned with visual features to compute a difficulty score, enabling efficient coreset selection without training the downstream model on the full dataset. 
See App.~\ref{app:additional_rw} for discussion about other CS methods. 

% Another line of work leverages {\em geometric or diversity-based criteria}, such as k-center greedy selection~\cite{sener2017active} or clustering-based methods~\cite{feldman2020turning}, to ensure coverage of the data distribution. Although computationally more efficient than training-dynamics-based approaches, these methods often rely on pairwise distance computations, which can become prohibitive at scale and may not capture semantic importance of samples.  
% More recently, {\em gradient- and influence-based methods}~\cite{koh2017understanding,killamsetty2021gradmatch} estimate the contribution of each sample to model training by approximating its effect on the loss or gradients. While theoretically grounded, these methods typically involve higher-order computations or repeated optimization, limiting their scalability.  

% Coreset Selection (CS) aims to identify a representative subset of a dataset such that models trained on this subset achieve performance comparable to training on the full dataset. Unlike dataset distillation, which synthesizes new data, coreset methods operate directly on the original dataset through data pruning. Existing CS approaches can be broadly categorized based on how they estimate the importance of samples.

% In this work, we select two state-of-the-art coreset selection methods to compare and benchmark against dataset distillation methods.

\paragraph{Reproducibility and reality checks in ML.}
A growing body of work shows that reported progress in ML can depend on evaluation protocols, undocumented design choices, and baseline selection.
\cite{gundersen2018state} found that most published ML studies lack the documentation needed for reproduction, and \cite{gundersen2022sources} categorized sources of irreproducibility into study design, implementation, and evaluation.
\cite{raff2019step} attempted reproductions of various ML papers and identified factors that predict reproducibility, while ~\cite{belz2021systematic} report that reproductions often failed to recover the originally reported results.
% Similar ``reality checks'' in metric learning~\cite{musgrave2020metric} and network pruning~\cite{blalock2020state} showed that apparent gains shrink or vanish under fair, standardized comparisons.
Closely related to our setting, Lu et al.~\cite{lu2023coreset} found that for logistic regression many sophisticated sampling methods fail to outperform random sampling. %, a simple baseline frequently omitted from comparisons.
We observe an analogous pattern in DD i.e., several SOTA DD methods fail to outperform random subsets (Sec.~\ref{sec:experiments}), despite being substantially more expensive to construct.
% However, unlike their setting, prior work~\cite{zheng2022coverage} has shown that AUM scores paired with stratified sampling produce coresets that outperform random sampling.  
This suggests that the conclusions about the superiority of DD depend on the evaluation protocol and on the omission of stronger/scalable CS methods~\cite{zheng2022coverage,mehra2025coreset}.

{\bf Prior works on rethinking evaluation protocols for DD.}
% Recent works have studied the evaluation practices in DD. 
% \cite{qin2024label} first showed that soft labels were the main driver of recent DD performance gains implying that the particular images matter less than the label information.
% \cite{li2025dd} assembled DD-Ranking, and reported that inconsistent training tricks (such as soft labels, data augmentation) had inflated the effectiveness of DD methods. 
% Our experiments strongly corroborate their finding that with abundant label knowledge, the performance gap between distilled and real subsets can vanish, reinforcing the importance of hard-label evaluation to measure any intrinsic merits of DD.
% Concurrent to our work, \cite{zhong2025rectified, dey2026rethinking,xiao2025rethinking} proposed to compare DD methods against CS methods.
% These new results extend the above works by refuting the notion that improved synthetic images have overtaken coreset selection, and they highlight that incorporating robust coreset baselines into evaluation of new DD methods is essential to avoid misinterpreting incremental improvements as fundamental breakthroughs.
% Recent works have begun to rethink the way DD methods are evaluated. 
Within DD specifically, Qin et al. \cite{qin2024label} showed that soft labels are the primary driver of DD gains, implying that the particular images often matter less than the label information. 
Li et al. \cite{li2025dd} assembled DD-Ranking to re-evaluate prior DD methods and reported that inconsistent training tricks (such as soft labels and data augmentation) had inflated the apparent effectiveness of DD. 
Our experiments strongly corroborate these insights, i.e., with rich label knowledge the performance gap between distilled and real subsets can vanish, underscoring the importance of hard-label evaluation to measure the merits of DD. 
% In other words, we confirm that dense teacher supervision can mask differences in dataset quality, making it essential to standardize evaluation protocols.

Concurrent with our work, \cite{zhong2025rectified, dey2026rethinking,xiao2025rethinking} have proposed unified benchmarks comparing DD and CS.
\cite{zhong2025rectified} introduce Rectified Decoupled Dataset Distillation (RD$^3$), focusing on decoupled DD methods, and \cite{dey2026rethinking, xiao2025rethinking} concurrently examine large-scale DD vs. CS, reinforcing that no existing DD method convincingly outperforms strong CS in hard-label settings. 
% Notably, as of writing, these concurrent works have not released their distilled datasets limiting direct SOTA-versus-SOTA comparisons. 
In contrast, our work provides a comprehensive evaluation of current SOTA DD vs. SOTA CS methods under uniform conditions. 
We extend prior analyses by showing that even recently proposed DD methods struggle to beat optimized real data subsets when hard labels are used, often by a large margin. % than previously observed (due to persistent diversity limitations in synthetic data). 
Moreover, we evaluate DD and CS along additional dimensions such as representativeness, diversity, and quality of images and show the CS methods perform better on all of these metrics. 
Thus, our results highlight that strong CS baselines must be included when evaluating new DD methods to avoid misinterpreting the effectiveness of DD methods for data efficiency use cases.
\section{Experimental Design for Evaluating Distilled Sets and Coresets}
\begin{figure}
% \begin{wrapfigure}{r}{0.6\linewidth}
\centering{\includegraphics[width=.85\textwidth]{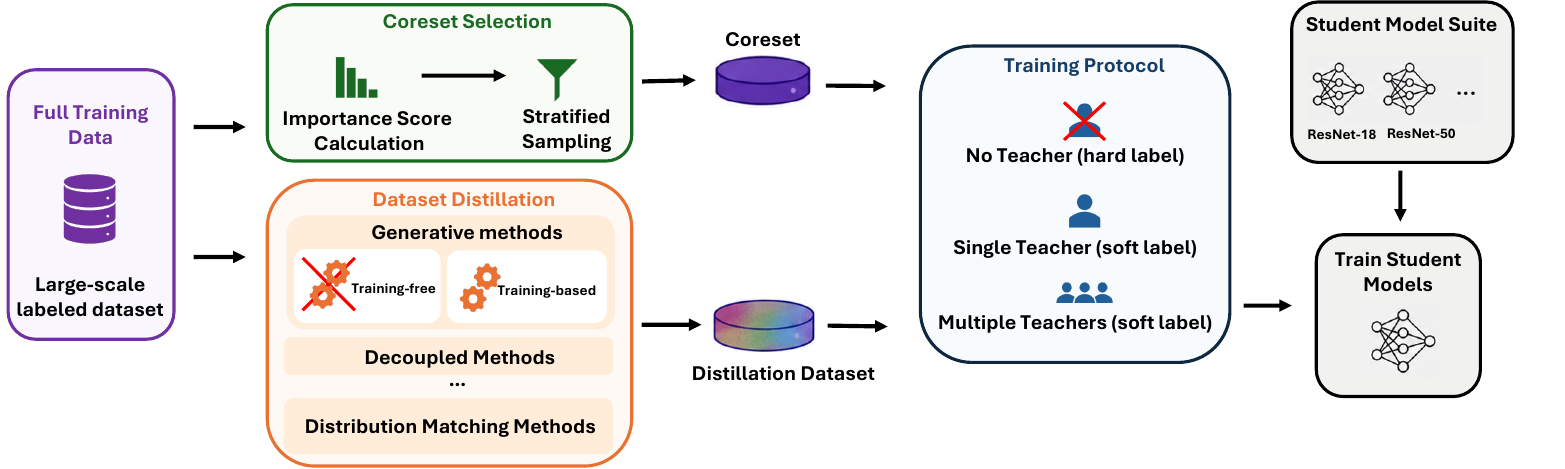}
  % \vspace{-0.95cm}
  \caption{\textbf{Experiment Design Overview:} Using the full dataset, we create coresets using CS methods~\textit{(green)} and distilled datasets using DD methods~\textit{(orange)}. We then train various student models on these coresets or distilled datasets via three training protocols and benchmark the performance of the trained student models on the test set. We experiment with three training protocols~\textit{(blue)}; using hard labels without a teacher, and using soft labels from single and multiple teachers.} 
  % and using soft labels from multiple teachers.  }}
  \label{fig:overview-fig}
  }
  % \vspace{-1.4cm}
% \end{wrapfigure}
\end{figure}

In this Section, we discuss details of our experiment design including the methods we used for CS and DD along with the evaluation protocols. See Fig.~\ref{fig:overview-fig} for an overview.

\subsection{DD Approaches}
% In this work, we compare state-of-the-art methods from each of these categories to understand dataset distillation better. 
Although a wide range of methods have been proposed for DD, we compared seven recent and representative DD methods that collectively span a wide range of design choices and assumptions in large-scale distillation (as can also be seen from Table~\ref{tab:related_methods}). 
Specifically, we include VLCP~\cite{zou2025vlcp}, D3HR~\cite{zhao2025taming}, and Minimax~\cite{gu2024efficient}, which are \emph{training-based generative} methods that train or fine-tune a generative model to synthesize the distilled dataset.
We also evaluate DiT-Distillation~\cite{jin2024fast}, MGD~\cite{chan2025mgd3}, and ManifoldGD~\cite{roy2026manifoldgd}, which are \emph{training-free generative} methods that leverage pretrained generative models and construct distilled datasets via guided sampling. Specifically, MGD and ManifoldGD perform guidance in the reverse diffusion process to steer samples toward more informative regions.
Lastly, we include FADRM+~\cite{cui2026fadrm}, a recent \emph{decoupled distillation} method based on the SRe2L \cite{yin2023squeeze} pipeline that optimizes samples to recover global and local information via data residual matching. % dataset construction from model training.
These SOTA methods were released recently with access to code/distilled data, and cover a wide spectrum of DD methods, enabling a comprehensive comparison.

\subsection{CS Approaches}

While many methods have been proposed for CS, we primarily focus on geometry and score‑based approaches due to their efficiency of computation and effectiveness as shown in prior works \cite{pleiss2020identifying,toneva2018empirical,zheng2022coverage,mehra2025coreset}. 
These methods leverage the geometry of the data or the training dynamics of a model trained on the full dataset $\mathcal{D}$ to estimate the importance of each training sample and then use a sampling strategy to form the coreset.
In this work, we focus on two training dynamics based approaches based on accumulated margin (AUM) \cite{pleiss2020identifying}, and concept‑guided importance scoring \cite{mehra2025coreset}. Along with this we use simple geometry-based approach based on clustering.
% While a wide range of methods have been proposed for coreset selection, we primarily focus on score‑based approaches that leverage the training dynamics of a model trained on the full dataset $\mathcal{D}$ to estimate the importance of each training sample. Recent studies have shown that training‑dynamics–based scoring methods consistently outperform heuristic or geometry‑based alternatives for coreset construction. In this work, we focus on three such approaches based on accumulated margin (AUM), forgetting events, and concept‑guided importance scoring.
In addition to these CS methods, we also include uniform random sampling (done class wise) which is a widely used baseline in CS.

For the clustering-based approach, we start by extracting DINO \cite{caron2021emerging} features, and then cluster features of each class into 10 clusters. Based on these, we assign each sample an importance score as its cosine distance to its nearest cluster centroid.
For the AUM-based approach \cite{pleiss2020identifying}, we compute the importance of each training sample based on classifier's margin during training. 
Specifically, given a model $f_{\theta_\mathcal{D}}$ trained on $\mathcal{D}$, the margin of a sample $(x,y)$ at epoch $t$ is defined as
$M^t(x,y) = f_{\theta_\mathcal{D},y}^t(x) - \max_{y' \neq y} f_{\theta_\mathcal{D},y'}^t(x)$,
where $f_{\theta_\mathcal{D},y'}^t(x)$ denotes the prediction logit (or likelihood) for class $y'$ at epoch $t$.
The \emph{area under the margin (AUM)} score after $T$ epochs of training is then computed as
% \begin{equation}
\(\mathrm{AUM}(x,y) = \frac{1}{T} \sum_{t=1}^{T} M^t(x,y).\)
% \end{equation}
% Prior work has shown that AUM effectively captures sample difficulty and informativeness, and can be used to identify mislabeled examples, forgetting events, and representative training samples for coreset selection
% \cite{pleiss2020identifying, zheng2022coverage, zheng2024elfs}.
% Forgetting-based methods quantify the number of times a training example transitions from being correctly classified to misclassified during training. Intuitively, frequently forgotten examples indicate instability in the training signal. In our implementation, we additionally assign a large forgetting score to samples that are \emph{never} correctly classified throughout training, treating them as consistently difficult examples.
Since training a downstream model on the full dataset could be inefficient, \cite{mehra2025coreset} proposed an alternative that leverages LLM-based concept bottleneck models to make CS even more efficient.
% In addition to training-dynamics based scoring, we consider a \emph{concept-guided} coreset selection strategy that incorporates semantic structure via class-level concepts, following prior work \cite{mehra2025coreset}. 
% This method uses semantic structure via class-level concepts extracted from pretrained LLMs and uses concept alignment to assess importance of each sample for CS. 
In this method, each image is embedded using a frozen visual encoder and compared against a set of LLM-generated concept embeddings to obtain a concept similarity representation.
A lightweight linear concept bottleneck classifier is trained on these representations for a small number of epochs, while recording the classification margin of each sample at every epoch.
The final \emph{concept-guided AUM} score, $\mathrm{AUM}_{\text{concept}}$, is computed by averaging these margins over the training epochs. 
% \(
% \mathrm{AUM}_{\text{concept}}(x,y) = \frac{1}{T} \sum_{t=1}^{T} \big( f_y^t(x) - \max_{y' \neq y} f_{y'}^t(x) \big).
% \)
See App.~\ref{app:additional_rw} for additional details of this method.

{\bf Sampling examples to form a coreset.} %\AM{CS has not been class-based}
A crucial step in CS is translating importance scores into an actual subset of training samples. 
Although retention of hard examples has often been shown to produce competitive performance, recent studies \cite{zheng2022coverage, sorscher2022beyond} demonstrate that this strategy can lead to catastrophic degradation of downstream performance when the coreset size is small, precisely the regime considered in this work, where IPC is limited. 
Such failures are largely attributed to poor distributional coverage and the prevalence of noisy or mislabeled samples among the hardest examples.
To mitigate these issues, we adopt \emph{Coverage‑centric Coreset Selection} (CCS) proposed by \cite{zheng2022coverage}, which explicitly balances sample difficulty with coverage of the data distribution. 
CCS first filters out a fraction of the hardest, potentially unreliable samples and then applies stratified sampling over the remaining difficulty scores to ensure uniform representation across easy and hard samples of the dataset. 
This strategy has been shown to consistently outperform both random sampling and naive hard‑sample selection across a wide range of coreset sizes \cite{zheng2022coverage,mehra2025coreset,zheng2024elfs}.
Since CS is not class-based we adapt CCS to operate under this constraint (see Alg.~\ref{alg:ipc_ccs} in App.~\ref{app:algo_ccs}).
Thus, rather than selecting a fixed fraction of the dataset, we construct the coreset independently for each class by selecting a fixed number of samples per class using CCS‑style sampling. 
This design ensures both strict class balance and diversity in sample difficulty within each class.

\subsection{Evaluation Methodology}
% As shown in Eqs.~\ref{eq:distillation} and~\ref{eq:coreset}, effectiveness of both DD and CS methods depends on how good the condensed sets perform in comparison to the model trained on original dataset. 
% In this section, we present three commonly used training and evaluation protocols that we use to assess the effectiveness of DD and CS for the same IPC setting.
% Evaluation of distilled datasets is a crucial component because performance depends not only on the synthetic data itself but also on the downstream training protocol. 
The standard evaluation pipeline adopted in
DD~\cite{zhao2023dataset,cui2026fadrm,shao2024generalizedlargescaledatacondensation} trains a model from scratch on the distilled set and evaluates it on a held-out real test set. 
However, recent DD works~\cite{Su_2024_CVPR,gu2024efficient,zhao2025taming,chan2025mgd3,roy2026manifoldgd} showed that this setup is highly sensitive to factors such as label type, augmentation strategy, and optimization details, especially in large scale setting, making evaluation methodologies a critical design choice in DD. 
% Below we detail the three popular training and evaluation methodologies adopted in DD works. 
To disentangle the effect of data quality and evaluation pipeline we evaluate distilled sets and coresets, using three evaluation protocols which are used in prior DD works. 

The first methodology uses the {\bf hard-label protocol} where each sample is assigned the class as provided in the distilled set or the coreset, mirroring standard supervised learning. 
This setup is closest to real-world deployment, depending only on the condensed set, and avoiding reliance on auxiliary supervision. 
% but it can underutilize the representational capacity of distilled samples. 
% The hard-label protocol is increasingly emphasized as a robust evaluation baseline in DD works, as they directly test whether distilled data captures class semantics without relying on any teacher-induced structure~\cite{zhao2023dataset,shao2024elucidating}.  
The next methodology uses the {\bf single teacher soft-label protocol}, implemented via a student--teacher (knowledge distillation) framework~\cite{hinton2015distilling}, which assigns each sample a probability distribution over classes produced by a pretrained teacher model (trained on the full dataset). 
This protocol is widely adopted in modern DD works~\cite{zou2025vlcp,li2025diff,gu2024efficient} because soft labels encode richer inter-class relationships and provide additional supervisory signal, leading to improved downstream performance, especially in large-scale settings.  
The third protocol is the {\bf multi-teacher soft-label protocol} \cite{shao2024elucidating} which extends the student--teacher paradigm by leveraging multiple teachers and hybrid supervision. 
In this setting, distilled data is optimized and evaluated using a combination of soft labels (from one or more teachers) and a hard-label constraints within a unified objective. 
This approach aims to balance the richness of soft supervision with the stability and generalization of hard labels, and have been shown to improve robustness across architectures and training settings~\cite{shao2024elucidating,shao2024generalizedlargescaledatacondensation}.  

% In this work, we evaluate distilled datasets under all three protocols to provide a comprehensive understanding of their quality and generalization behavior.

\subsection{Datasets, Models and Implementation Details}
\label{sec:implementation_details}
{\bf Datasets.}
We conduct experiments on large-scale image classification benchmarks commonly used in DD literature. 
Our primary evaluation is performed on the full ImageNet-1K dataset~\cite{deng2009imagenet}, which contains $1,000$ classes and serves as a challenging large-scale benchmark for both CS and DD methods. 
To study the effect of dataset scale and facilitate comparison with prior DD works, we also use two standard ImageNet subsets: ImageNet-100~\cite{tian2020contrastive}, consisting of $100$ classes sampled from ImageNet-1K, and ImageNette~\cite{howard2019imagenette}, a subset of $10$ easily classifiable ImageNet classes. 
% These subsets are widely used to evaluate the efficiency of data in reduced category complexity.

{\bf Model Architectures.}
% We evaluated all methods using model architectures commonly adopted in the dataset distillation literature. 
For all datasets, we use ResNet-18 and ResNet-50 as student models.  
For ImageNet-100 and ImageNette, we additionally use ResNet-AP-10 model similar to prior works. Additional evaluation with ViT-B/16 and Swin are evaluated in App.~\ref{app:architecture_generalization}

{\bf Implementation Details.}  We use the official codes from the GitHub repositories (see App.~\ref{app:implementation_details}) of various prior works for obtaining distilled datasets, generating coresets, and evaluation.
We evaluated all CS and DD approaches with 10 and 50 IPC similar to prior works. 
 % for all our experiments~(Table \ref{tab:imagenet1k-results}-Table \ref{tab:misratio_ablation})..
We briefly describe the details of various evaluation protocols here.  
For the \textit{hard label evaluation protocol}, we use the training protocol used in~\cite{roy2026manifoldgd,chan2025mgd3}. 
In this protocol, we generate the condensed set and use the assigned class labels to train an image classifier from scratch. 
%, and evaluates the network on the original test set.  
We apply random resize-crop (224 $\times$ 224) and CutMix during training. 
%are applied as augmentation techniques during the student model’s training. 
The model is trained on condensed sets for $300$ epochs with a batch size of $64$ 
% and a learning rate of 5E-4 
for both IPCs for ImageNet-1K; $1300$ and $1000$ epochs, respectively, for IPC $10$ and $50$ for ImageNet-100 and $2000$ and $1500$ epochs for IPC $10$ and $50$ for ImageNette dataset.
We use Stochastic Gradient Descent (SGD) as the optimizer with cross-entropy loss and a learning rate of $0.01$ with a weight decay of 5E-4. 
We use a learning rate decay scheduler at the $2/3$ and $5/6$ points of the training process, with the decay factor ($\gamma$) set to $0.2$. %Cross-entropy was used as the loss objective.  

For the other two protocols, we follow two evaluation pipelines proposed in RDED~\cite{sun2024diversityrealismdistilleddataset} and EDC~\cite{shao2024elucidating}. 
In RDED's evaluation method, a single teacher model is used whereas in EDC's evaluation method multiple teacher models are used for obtaining soft labels as mentioned below. 
As proposed in~\cite{sun2024diversityrealismdistilleddataset}, region-based soft-labels are generated with a pretrained teacher network(s) as follows:
\(
y_{i,m} = \phi_{\mathcal{T}}(x_{i,m}),
\)
where $\phi_{\mathcal{T}}$ is the pretrained model (ResNet-18 in this case) and $x_{i,m}$ is the $m$-th crop of the $i$-th image. 
Then a student model $\phi_{\mathcal{S}}$ is trained on the condensed sets by minimizing the loss computed as 
\(
\mathcal{L} = - \sum_{i} \sum_{m} y_{i,m} \log \phi_{\mathcal{S}}(x_{i,m}).
\) 
% We follow the same process for all three datasets, ImageNet-1k, ImageNet-100 and ImageNette dataset for both IPC 10 and 50. 
% We use ResNet-18 as the teacher model and experiment with ResNet-18 and ResNet-50 as the student models for this setup. 
 %, ImageNet-1k, ImageNet-100 and ImageNette dataset.
In EDC's evaluation protocol, we use four teacher models, namely ResNet-18, MobileNet, EfficientNet, and ShuffleNet, pretrained on the respective full datasets to obtain teacher labels for a sample denoted as $y_t$.  
The student model is then trained from scratch by minimizing the MSE loss between the student labels, $y_s$ along with cross-entropy loss with assigned one-hot label $y_{onehot}$ of the sample in the condensed set. The total loss is denoted as 
\(
\ell(x, y_{onehot}) = \ell_{CE}(y_{onehot}, y_s) + \kappa \ell_{MSE}(y_t, y_s),
\)
where $\kappa$ is set to 0.025. 
We train for $300$ epochs for both IPC values and for all three datasets.
For each evaluation, we train classifiers from \emph{scratch} on the condensed sets three times to get the accuracy, and the mean and standard deviation on the test set are reported in all tables.

To generate the AUM coresets, for 10 and 50 IPC, we train a ResNet-18 model from scratch on the respective dataset for $60$ epochs and record the $\mathrm{AUM}$ scores for the entire dataset.
For concept-based AUM, the classwise concepts are extracted using the LLaVa model to form the concept bottleneck and then CLIP ViT-B32 model is used to assess similarity between visual and concept features. 
We then train the single linear layer, minimizing cross entropy loss for 100 epochs and record $\mathrm{AUM}_{\text{concept}}$ for all the samples.
We then use modified stratified sampling proposed in Alg.~\ref{alg:ipc_ccs} to generate the coreset for all CS methods. We note that the same coreset is used for all student architectures, i.e., for AUM we use the coreset formed using training dynamics for ResNet-18, even when the student model is ResNet-50. This is done to highlight the effectiveness of coresets independent of the student model.

\section{Results and Analysis}
\label{sec:experiments}
% Here, w
We evaluate CS and DD methods under standardized datasets, architectures, and evaluation protocols. Due to space limitation, some experiments and ablations have been deferred to the App.~\ref{app:additional_results}. 
% Our experimental design emphasizes controlled comparisons across multiple dataset scales and evaluation pipelines, enabling a fair assessment of the practical trade-offs between these two approaches. 
% All experiments in this section were conducted with a single NVIDIA A100 GPU. 
% % Due to space constraints, detailed hyperparameter settings, additional ablation studies, and visualizations of synthesized datasets are provided in the Appendix.
% {\bf Dataset Distillation and Coreset Selection Baselines. } As summarized in Section~\ref{}, we choose \textit{seven dataset distillation methods}~(VLCP~\cite{zou2025vlcp}, D3HR~\cite{zhao2025taming}, Minimax~\cite{gu2024efficient}, DiT Distillation~\cite{}, MGD~\cite{chan2025mgd3}, and ManifoldGD~\cite{roy2026manifoldgd}) and \textit{three coreset selection methods}~(Random, AUM~\cite{} and Concept-Based CS~\cite{mehra2025coreset}). 

\subsection{Main Results and Discussion}
{\bf Results on ImageNet-1K.}
Table~\ref{tab:imagenet1k-results} summarizes accuracy on ImageNet‑1K under three evaluation protocols, across two student architectures (ResNet‑18 and ResNet‑50).
Under the hard‑label protocol, which does not rely on any teacher model, CS consistently matches or outperforms SOTA DD methods, with the effect being most pronounced at 10 IPC.
For example, with ResNet‑18 at 10 IPC, AUM achieves a 3.3\% gain over the best performing DD method. 
A similar improvement of 4.8\% is observed with ResNet-50. 
These gains persist at 50 IPC as well where AUM remains the best method, exceeding the strongest DD method by more than 4\%.
This demonstrates that best CS method surpasses best DD method under this challenging and realistic protocol. 

With soft labels from a single teacher (RDED \cite{sun2024diversityrealismdistilleddataset}), while the overall accuracy improves across all methods, CS methods continue to perform competitively to distilled data. 
For 10 IPC, coreset methods exceed the best distillation results for both student models, while at 50 IPC, Concepts surpasses all DD methods for ResNet-18 (61.0\% vs. 60.4\%), while Clustering achieves the best performance for ResNet-50 (66.2\%).
% at 50 IPC, Concepts surpass all distillation methods for ResNet‑18 by 1\%, while AUM and Concepts jointly achieve the best performance for ResNet‑50 (66.0\%).
These results indicate that CS methods improve performance over DD methods when teacher supervision commonly used in DD works is available. 
% distillation‑friendly supervision is available, and can fully exploit soft guidance without relying on generative synthesis.
Under the EDC protocol, which aggregates the supervision of multiple different teachers, CS methods are consistently better than DD methods. 
For 10 IPC, AUM yields the highest accuracy while at 50 IPC, Clustering achieves the best results across both students. 

Lastly, under all three evaluation protocols, many DD methods underperform the simplest CS baseline of random subsets highlighting the limitations in the practical success of current DD approaches. 
% \AM{mention comparison to random}

% 18.9%, outperforming the best distillation baseline (ManifoldGD at 15.6%) by +3.3 points. A similar trend holds for ResNet‑50, where AUM improves over the strongest distillation method by +4.8 points at IPC = 10.
% This effect is most pronounced at low IPC (10), where the discrepancy between synthetic data quality and real‑label supervision is most challenging.
\begin{table}[t]
\centering
\caption{
Comparison of accuracy (\%, mean±std) of condensed sets on \textbf{ImageNet‑1K}. 
CS methods, particularly AUM, consistently match or outperform DD methods under hard‑label evaluation and remain competitive under soft‑label protocols. (Best numbers for both DD and CS are highlighted)
% In contrast, the best‑performing distillation method varies across evaluation pipelines, highlighting strong protocol dependence for DD methods on large‑scale datasets.
%\textbf{Experimental Results for ImageNet-1K. }We report top-1 accuracy (\%, mean and standard deviation) for two student models~(ResNet-18 and ResNet-50) and two IPCs (10 and 50) with three evaluation pipelines, hard-label protocol~(requires no teacher model), soft-label protocol~(RDED) with a single teacher model~(ResNet-18), and finally  soft-label protocol (EDC) that uses multiple teacher models~(ResNet-18, MobileNet\_v2, ShuffleNet\_v2\_x0\_5, AlexNet). 
% Distillation methods are listed first, followed by coreset selection baselines. 
}
\label{tab:imagenet1k-results}
{\renewcommand{\arraystretch}{1.35}
\resizebox{0.99\textwidth}{!}{
\begin{tabular}{rcccccccc@{\hspace{1.5cm}}cccc}
\toprule[1.25pt]

 & &
\multicolumn{7}{c}{\cellcolor{Cerulean!30}{\textbf{{\large Distillation Methods}}}} 
& \multicolumn{4}{c}{\cellcolor{RubineRed!20}{\textbf{{\large Coreset Selection Methods}}}} \\

\cmidrule{3-9}\cmidrule{10-13}

 & &
\makecell{\textbf{VLCP} \cite{zou2025vlcp}} & \makecell{\textbf{D3HR} \cite{zhao2025taming}} & \makecell{\textbf{Minimax} \cite{gu2024efficient}} & \makecell{\textbf{DiT} \cite{peebles2023scalable}} & \makecell{\textbf{MGD}\cite{chan2025mgd3}} & \makecell{\textbf{ManifoldGD} \cite{roy2026manifoldgd}} & \makecell{\textbf{FADRM+} \cite{cui2026fadrm}} &
\makecell{\textbf{Random}} & \makecell{\textbf{Clustering}} & \makecell{\textbf{AUM} \cite{pleiss2020identifying}} &\makecell{\textbf{Concepts} \cite{mehra2025coreset}} \\

\midrule

% \rowcolor{gray!40}  \multicolumn{12}{c}{\textsc{\textbf{\large {Evaluation Protocol: Hard Label}}}} \\

\rowcolor{gray!20}\textbf{Student Model} & \textbf{IPC} & \multicolumn{11}{c}{\textsc{\textbf{\large {Evaluation Protocol: Hard label}}}}\\
\cmidrule{1-13}

\multirow{2}{*}{\textbf{ResNet-18}} & \textbf{10} &
\mstd{14.6}{0.4} & \mstd{14.1}{0.3} & \mstd{14.5}{0.1} & \mstd{12.9}{0.1} &
\mstd{13.5}{0.4} & \textbf{\mstd{15.6}{0.2}} & \mstd{9.6}{0.1} &
\mstd{9.3}{0.2} & \mstd{16.7}{0.4} & \textbf{\mstd{\cellcolor{green!20}{18.9}}{0.4}} & \mstd{15.0}{0.2} \\

& \textbf{50} &
\mstd{34.9}{0.2} & \mstd{31.2}{0.1} & \mstd{32.5}{0.1} & \mstd{31.2}{0.1} &
\mstd{39.2}{0.2} & \textbf{\mstd{39.3}{0.0}} & \mstd{18.6}{0.4} &
\mstd{39.2}{0.2} & \mstd{42.5}{0.3} & \textbf{\mstd{\cellcolor{green!20}{43.5}}{0.1}} & \mstd{42.6}{0.1} \\

\cmidrule{2-13}

\multirow{2}{*}{\textbf{ResNet-50}} & \textbf{10} &
\mstd{10.4}{2.6} & \mstd{10.2}{2.2} & \mstd{12.3}{0.8} & \textbf{\mstd{12.5}{0.4}} &
\mstd{8.6}{1.4} & \mstd{9.2}{4.0} & \mstd{6.3}{4.0} &
\mstd{6.5}{0.12} & \mstd{14.6}{1.4} & \textbf{\mstd{\cellcolor{green!20}{17.3}}{1.1}} & \mstd{12.8}{1.3} \\

& \textbf{50} &
\mstd{39.0}{0.2} & \mstd{35.7}{0.1} & \mstd{37.1}{0.4} & \mstd{30.8}{0.1} &
\textbf{\mstd{44.8}{0.4}} & \mstd{44.4}{0.3} & \mstd{19.6}{0.1} &
\mstd{44.1}{0.1} & \mstd{47.5}{0.3} & \textbf{\mstd{\cellcolor{green!20}{48.9}}{0.3}} & \mstd{48.3}{0.4} \\

\midrule

% \rowcolor{gray!40}  \multicolumn{12}{c}{\textsc{\textbf{\large {Evaluation Protocol: Soft Label (RDED)}}}} \\

\rowcolor{gray!20}\textbf{Student Model} & \textbf{IPC} & \multicolumn{11}{c}{\textsc{\textbf{\large {Evaluation Protocol: Single teacher soft label (RDED)}}}}\\
\cmidrule{1-13}

\multirow{2}{*}{\textbf{ResNet-18}} & \textbf{10} &
\mstd{41.8}{0.2} & \mstd{42.8}{0.3} & \mstd{44.1}{0.3} & \mstd{41.9}{0.3} &
\mstd{45.9}{0.4} & \textbf{\mstd{46.0}{0.2}} & \textbf{\mstd{46.0}{0.3}} &
\mstd{44.6}{0.6} & \mstd{46.5}{0.4} & \textbf{\mstd{\cellcolor{green!20}{46.9}}{0.4}} & \mstd{46.4}{0.2} \\

& \textbf{50} &
\mstd{58.9}{0.1} & \mstd{58.8}{0.3} & \mstd{58.7}{0.2} & \mstd{58.5}{0.1} &
\mstd{60.3}{0.1} & \textbf{\mstd{60.4}{0.1}} & \mstd{58.5}{0.2} &
\mstd{60.6}{0.1} & \mstd{60.9}{0.2} & \mstd{60.9}{0.1} & \textbf{\mstd{\cellcolor{green!20}{61.0}}{0.1}} \\

\cmidrule{2-13}

\multirow{2}{*}{\textbf{ResNet-50}} & \textbf{10} &
\mstd{47.4}{0.4} & \mstd{47.9}{0.9} & \mstd{49.8}{0.8} & \mstd{47.6}{2.0} &
\mstd{49.2}{0.6} & \mstd{49.8}{0.8} & \textbf{\mstd{50.7}{0.3}} &
\mstd{49.9}{0.6} & \mstd{50.4}{1.9} & \mstd{51.5}{1.6} & \textbf{\mstd{\cellcolor{green!20}{52.0}}{1.5}} \\

& \textbf{50} &
\mstd{64.8}{0.1} & \mstd{64.6}{0.1} & \mstd{64.3}{0.1} & \mstd{64.3}{0.2} &
\mstd{65.3}{0.2} & \textbf{\mstd{65.6}{0.2}} & \mstd{63.7}{0.1} &
\mstd{65.9}{0.14} & \textbf{\mstd{\cellcolor{green!20}{66.2}}{0.2}} & \mstd{66.0}{0.1} & \mstd{66.0}{0.1} \\

\midrule

% \rowcolor{gray!40}  \multicolumn{12}{c}{\textsc{\textbf{\large {Evaluation Protocol: Soft Label (EDC)}}}} \\

\rowcolor{gray!20}\textbf{Student Model} & \textbf{IPC} & \multicolumn{11}{c}{\textsc{\textbf{\large {Evaluation Protocol: Multiple teachers soft label (EDC)}}}}\\
\cmidrule{1-13}

\multirow{2}{*}{\textbf{ResNet-18}} & \textbf{10} &
\mstd{47.5}{0.6} & \mstd{47.7}{0.2} & \mstd{49.2}{0.0} & \mstd{48.3}{0.3} &
\mstd{50.1}{0.2} & \textbf{\mstd{51.2}{0.3}} & \mstd{48.9}{0.1} &
\mstd{50.8}{0.1} & \mstd{51.2}{0.3} & \textbf{\mstd{\cellcolor{green!20}{51.4}}{0.2}} & \mstd{51.2}{0.3} \\

& \textbf{50} &
\mstd{58.3}{0.2} & \mstd{57.7}{0.1} & \mstd{57.9}{0.0} & \mstd{57.4}{0.1} &
\mstd{59.5}{0.2} & \textbf{\mstd{59.8}{0.1}} & \mstd{56.6}{0.0} &
\mstd{60.5}{0.0} & \textbf{\mstd{\cellcolor{green!20}{60.7}}{0.1}} & \mstd{60.5}{0.3} & \mstd{60.6}{0.1} \\

\cmidrule{2-13}

\multirow{2}{*}{\textbf{ResNet-50}} & \textbf{10} &
\mstd{53.4}{0.2} & \mstd{53.8}{0.4} & \mstd{54.5}{0.6} & \mstd{53.8}{0.4} &
\mstd{55.4}{0.6} & \textbf{\mstd{56.0}{0.5}} & \mstd{54.2}{0.5} &
\mstd{56.6}{0.7} & \mstd{56.5}{0.4} & \textbf{\mstd{\cellcolor{green!20}{56.7}}{1.0}} & \mstd{56.4}{0.7} \\

& \textbf{50} &
\mstd{64.2}{0.1} & \mstd{63.7}{0.2} & \mstd{63.7}{0.1} & \mstd{63.5}{0.1} &
\textbf{\mstd{65.1}{0.2}} & \textbf{\mstd{65.1}{0.1}} & \mstd{62.7}{0.1} &
\textbf{\mstd{\cellcolor{green!20}{66.2}}{0.0}}& \mstd{66.1}{0.1} & \mstd{65.8}{0.2} & \mstd{65.1}{0.1} \\

\bottomrule[1.25pt]
\end{tabular}
}
}
% \vspace{-20pt}
\vspace{-10pt}
\end{table}

{\bf Results on ImageNet-100.}
\begin{table}[t]
\centering
\caption{Comparison of accuracy (\%, mean±std) of condensed sets on \textbf{ImageNet‑100}. Across protocols, CS methods perform competitively to DD methods, outperforming them significantly under the hard‑label protocol. 
% Experimental Results for ImageNet-100.} We report top-1 accuracy (\%, mean and standard deviation) for three student models~(ResNetAP-10, ResNet-18 and ResNet-50) and two IPCs (10 and 50) with three evaluation pipelines, hard-label protocol~(requires no teacher model), soft-label protocol~(RDED) with a single teacher model~(ResNet-18), and finally soft-label protocol (EDC) that uses multiple teacher models~(ResNet-18, MobileNet\_v2, ShuffleNet\_v2\_x0\_5, AlexNet). Distillation methods are listed first, followed by coreset selection baselines. 
}
\label{tab:imagenet100-results}
{\renewcommand{\arraystretch}{1.25}
\resizebox{0.99\textwidth}{!}{
\begin{tabular}{rcccccccc@{\hspace{1.5cm}}cccc}
\toprule[1.25pt]

 & &
\multicolumn{7}{c}{\cellcolor{Cerulean!30}{\textbf{{\large Distillation Methods}}}} 
& \multicolumn{4}{c}{\cellcolor{RubineRed!20}{\textbf{{\large Coreset Selection Methods}}}} \\

\cmidrule{3-9}\cmidrule{10-13}

 & &
\makecell{\textbf{VLCP} \cite{zou2025vlcp}} & \makecell{\textbf{D3HR} \cite{zhao2025taming}} & \makecell{\textbf{Minimax} \cite{gu2024efficient}} & \makecell{\textbf{DiT} \cite{peebles2023scalable}} & \makecell{\textbf{MGD}\cite{chan2025mgd3}} & \makecell{\textbf{ManifoldGD} \cite{roy2026manifoldgd}} & \makecell{\textbf{{FADRM+}} \cite{cui2026fadrm}} &
\makecell{\textbf{Random} } & \makecell{\textbf{Clustering} } & \makecell{\textbf{AUM} \cite{pleiss2020identifying}} &\makecell{\textbf{Concepts} \cite{mehra2025coreset}} \\

\midrule

% \rowcolor{gray!40}  \multicolumn{12}{l}{\textsc{\textbf{\large {Evaluation Protocol: Hard Label}}}} \\

\rowcolor{gray!20}\textbf{Student Model} & \textbf{IPC} & \multicolumn{11}{c}{\textsc{\textbf{\large {Evaluation Protocol: Hard label}}}}\\
\cmidrule{1-13}

\multirow{2}{*}{\textbf{ResNetAP-10}} & \textbf{10} &
\mstdgray{24.9}{0.4} & \textbf{\mstdgray{25.8}{0.1}} & \mstdgray{25.6}{0.1} & \mstd{23.6}{0.4} &
\mstd{23.1}{0.2} & \mstd{25.1}{0.4} & \mstdgray{16.9}{0.2} &
\mstd{18.4}{0.4} & \mstd{23.7}{0.3} & \textbf{\mstd{\cellcolor{green!20}{27.8}}{0.2}} & \mstd{23.0}{0.5} \\

& \textbf{50} &
\mstdgray{44.5}{0.5} & \mstdgray{43.8}{0.1} & \mstdgray{44.1}{0.7} & \mstd{37.2}{0.2} &
\mstd{46.4}{0.2} & \textbf{\mstd{48.0}{0.5}} & \mstdgray{31.7}{0.2} &
\mstd{40.3}{0.9} & \mstd{48.1}{0.2} & \textbf{\mstd{\cellcolor{green!20}{50.6}}{0.1}} & \mstd{47.2}{0.8} \\

\midrule

\multirow{2}{*}{\textbf{ResNet-18}} & \textbf{10} &
\mstdgray{23.3}{0.1} & \textbf{\mstdgray{24.7}{0.7}} & \mstdgray{24.2}{0.2} & \mstd{21.8}{0.2} &
\mstd{21.9}{0.4} & \mstd{23.1}{0.4} & \mstdgray{16.0}{0.3} &
\mstd{16.1}{0.7} & \mstd{23.9}{0.2} & \textbf{\mstd{\cellcolor{green!20}{28.1}}{0.7}} & \mstd{20.4}{0.8} \\

& \textbf{50} &
\mstdgray{46.7}{0.2} & \mstdgray{45.7}{0.2} & \mstdgray{47.1}{0.3} & \mstd{38.9}{0.1} &
\mstd{48.0}{0.5} & \textbf{\mstd{49.5}{0.1}} & \mstdgray{32.7}{0.4} &
\mstd{43.5}{1.0} & \mstd{53.0}{3.4} &  \textbf{\mstd{\cellcolor{green!20}{53.0}}{0.2}} & \mstd{47.2}{0.8} \\

\midrule

\multirow{2}{*}{\textbf{ResNet-50}} & \textbf{10} &
\mstdgray{17.5}{0.3} & \textbf{\mstdgray{20.0}{0.8}} & \mstdgray{18.2}{0.9} & \mstd{18.8}{0.3} &
\mstd{15.5}{1.1} & \mstd{16.5}{0.5} & \mstdgray{12.0}{0.3} &
\mstd{10.9}{0.9} & \mstd{18.7}{0.1} & \textbf{\mstd{\cellcolor{green!20}{23.5}}{0.5}} & \mstd{14.5}{1.1} \\

& \textbf{50} &
\mstdgray{46.9}{0.1} & \mstdgray{45.0}{0.7} & \mstdgray{48.3}{0.3} & \mstd{38.8}{0.5} &
\mstd{50.0}{0.2} & \textbf{\mstd{50.5}{0.2}} & \mstdgray{31.7}{0.6} &
\mstd{43.7}{1.1} & \mstd{52.8}{1.1} & \mstd{54.1}{0.6} & \textbf{\mstd{\cellcolor{green!20}{54.6}}{3.0}} \\

\midrule

% \rowcolor{gray!40}  \multicolumn{12}{l}{\textsc{\textbf{\large {Evaluation Protocol: Soft Label (RDED)}}}} \\

\rowcolor{gray!20}\textbf{Student Model} & \textbf{IPC} & \multicolumn{11}{c}{\textsc{\textbf{\large {Evaluation Protocol: Single teacher soft label (RDED)}}}}\\
\cmidrule{1-13}

\multirow{2}{*}{\textbf{ResNet-18}} & \textbf{10} &
\mstdgray{25.8}{0.7} & \mstdgray{27.4}{0.6} & \mstdgray{27.9}{1.4} & \mstd{26.0}{0.7} &
\mstd{30.0}{1.0} & \mstd{30.1}{0.4} & \textbf{\mstdgray{30.2}{0.6}} &
\mstd{27.7}{1.3} & \mstd{31.1}{0.6} & \textbf{\mstd{\cellcolor{green!20}{32.9}}{0.2}} & \mstd{28.9}{0.4} \\

& \textbf{50} &
\mstdgray{61.8}{0.6} & \mstdgray{59.3}{0.4} & \mstdgray{61.3}{0.4} & \mstd{56.6}{0.8} &
\textbf{\mstd{63.8}{0.1}} & \mstd{59.3}{0.4} & \mstdgray{62.9}{0.3} &
\mstd{66.2}{0.2} & \textbf{\mstd{\cellcolor{green!20}{66.7}}{1.7}} & \mstd{66.0}{0.4}& \mstd{66.6}{0.3} \\

\midrule

\multirow{2}{*}{\textbf{ResNet-50}} & \textbf{10} &
\mstdgray{23.9}{1.0} & \mstdgray{26.8}{1.1} & \mstdgray{26.7}{0.5} & \mstd{24.7}{1.6} &
\mstd{27.4}{0.6} & \textbf{\mstd{28.4}{0.2}} & \mstdgray{27.5}{1.1} &
\mstd{25.7}{0.8} & \mstd{30.5}{1.5} & \textbf{\mstd{\cellcolor{green!20}{30.6}}{0.6}} & \mstd{28.1}{0.8} \\

& \textbf{50} &
\mstdgray{64.3}{0.5} & \mstdgray{62.0}{0.5} & \mstdgray{63.2}{0.8} & \mstd{59.6}{0.4} &
\mstd{66.2}{1.5} & \textbf{\mstd{66.7}{1.0}} & \mstdgray{65.5}{0.6} &
\mstd{67.2}{0.8} & \textbf{\mstd{\cellcolor{green!20}{68.8}}{0.4}} & \mstd{67.7}{0.7} & \mstd{68.1}{0.1} \\

\midrule

% \rowcolor{gray!40}  \multicolumn{12}{l}{\textsc{\textbf{\large {Evaluation Protocol: Soft Label (EDC)}}}} \\

\rowcolor{gray!20}\textbf{Student Model} & \textbf{IPC} & \multicolumn{11}{c}{\textsc{\textbf{\large {Evaluation Protocol: Multiple teachers soft label (EDC)}}}}\\
\cmidrule{1-13}

\multirow{2}{*}{\textbf{ResNet-18}} & \textbf{10} &
\mstdgray{40.5}{0.4} & \mstdgray{45.1}{0.7} & \mstdgray{45.2}{0.9} & \mstd{42.1}{0.5} &
\mstd{43.7}{0.6} & \textbf{\mstd{46.33}{0.5}} & \mstdgray{42.9}{0.1} &
\mstd{40.8}{0.6} & \mstd{45.6}{0.1} & \textbf{\mstd{\cellcolor{green!20}{48.03}}{0.53}} & \mstd{44.2}{1.0} \\

& \textbf{50} &
\mstdgray{71.4}{0.7} & \mstdgray{70.8}{0.3} & \mstdgray{72.2}{0.4} & \mstd{69.0}{0.3} &
\mstd{73.4}{0.1} & \textbf{\mstd{74.03}{0.2}} & \mstdgray{69.6}{0.4} &
\mstd{73.3}{0.3} & \textbf{\mstd{\cellcolor{green!20}{74.3}}{0.3}} & \mstd{73.4}{0.5} & \mstd{73.9}{0.4} \\

\midrule

\multirow{2}{*}{\textbf{ResNet-50}} & \textbf{10} &
 \mstdgray{36.1}{0.5} & \mstdgray{41.0}{1.9} & \mstdgray{40.1}{1.3} & \mstd{38.2}{0.5} & \mstd{39.4}{0.3} & \textbf{\mstd{41.8}{1.2}} & \mstdgray{37.6}{1.0} & \mstd{37.1}{1.1} & \mstd{43.6}{1.2} & \textbf{\mstd{\cellcolor{green!20}{43.8}}{2.1}} & \mstd{39.1}{0.4} \\

& \textbf{50} &

\mstdgray{72.7}{0.9} & \mstdgray{71.7}{1.4} & \mstdgray{73.0}{1.1} & \mstd{69.3}{0.5} & \mstd{72.2}{1.7} & \textbf{\mstd{74.3}{0.8}} & \mstdgray{69.3}{0.5} & \mstd{73.6}{0.3} & \textbf{\mstd{\cellcolor{green!20}{74.6}}{1.0}} & \mstd{74.0}{0.4} & \mstd{73.1}{1.6} \\

\bottomrule[1.25pt]
\end{tabular}
}
}
%\vspace{-20pt}
\vspace{-15pt}
\end{table}
Table~\ref{tab:imagenet100-results} shows that CS methods remain highly competitive across all student architectures and evaluation protocols on this dataset. 
Under the hard‑label protocol, AUM substantially exceeds the performance of DD methods for all students and IPC. 
Even under the soft‑label protocols, CS methods match or outperform DD methods.
Specifically, in RDED, Clustering achieves the best results at 50 IPC, while AUM leads to the best results for 10 IPC. 
Under EDC protocol, CS methods are on par with DD for 50 IPC, but they perform better at 10 IPC. 
These results reinforce that CS provides a strong and robust alternative to DD. 
{\underline{Note.}} In Table~\ref{tab:imagenet100-results}, results of DD methods in gray color refer to cases where official distilled datasets for ImageNet‑100 were not available. % in the respective GitHub repositories. 
For such cases, we evaluate DD methods by restricting the available ImageNet‑1K distilled set to the subset of classes present in ImageNet‑100.
This choice was made primarily for training efficiency, since some methods (e.g., VLCP) require training a generative model for distillation on a new dataset which is computationally expensive.

{\bf Results on ImageNette.}
Table~\ref{tab:imagenette-results_1} shows that while DD methods are competitive with CS in the low IPC regime, they still suffer in the high IPC regime.
Specifically, under the hard-label protocol, the best DD method outperforms CS at 10 IPC for ResNetAP-10 and ResNet-18 and is on par with AUM for ResNet-50.
However, this advantage does not persist at 50 IPC, where Clustering outperforms DD methods. % for every student.
With soft labels from a single teacher (RDED), CS methods achieve the best accuracy in all settings, with even Random surpassing every DD method for ResNet-18/50 at 50 IPC.
Under the EDC protocol, DD methods lead at 10 IPC, while at 50 IPC CS methods achieve the best results.
This indicates that even on a small, easily separable dataset, the advantage of DD is limited to small budgets and specific evaluation protocols. %, and the results in Tables~\ref{tab:imagenet1k} and~\ref{tab:imagenet100} show that DD methods suffer further on larger-scale datasets, where gains from data-centric learning are most valuable.

\begin{table}[t]
\centering
\caption{
Comparison of accuracy (\%, mean$\pm$std) of condensed sets on \textbf{ImageNette}. %\AM{Trisha are the concept numbers in green correct?}
}
\label{tab:imagenette-results_1}
{\renewcommand{\arraystretch}{1.25}
\resizebox{0.99\textwidth}{!}{
\begin{tabular}{rcccccccc@{\hspace{1.5cm}}cccc}
\toprule[1.25pt]

 & & 
\multicolumn{7}{c}{\cellcolor{Cerulean!30}{\textbf{{\large Distillation Methods}}}} 
& \multicolumn{4}{c}{\cellcolor{RubineRed!20}{\textbf{{\large Coreset Selection Methods}}}} \\

\cmidrule{3-9}\cmidrule{10-13}

 & & 
\makecell{\textbf{VLCP} \cite{zou2025vlcp}} & \makecell{\textbf{D3HR} \cite{zhao2025taming}} & \makecell{\textbf{Minimax} \cite{gu2024efficient}} & \makecell{\textbf{DiT} \cite{peebles2023scalable}} & \makecell{\textbf{MGD}\cite{chan2025mgd3}} & \makecell{\textbf{ManifoldGD} \cite{roy2026manifoldgd}} & \makecell{\textbf{{FADRM+}} \cite{cui2026fadrm}} &
\makecell{\textbf{Random} } & \makecell{\textbf{Clustering} } &  \makecell{\textbf{AUM} \cite{pleiss2020identifying}} &\makecell{\textbf{Concepts} \cite{mehra2025coreset}} \\

\midrule

\rowcolor{gray!20}\textbf{Student Model} & \textbf{IPC} & \multicolumn{11}{c}{\textsc{\textbf{\large {Evaluation Protocol: Hard label}}}}\\
\cmidrule{1-13}

\multirow{2}{*}{\textbf{ResNetAP-10}} & \textbf{10} &
\mstd{61.9}{0.2} & \mstd{60.1}{0.2} & \mstd{57.3}{0.5} & \mstd{56.9}{0.5} &
\mstd{58.1}{0.3} & \textbf{\mstd{63.2}{0.5}} & \mstd{56.3}{0.2} &
\mstd{52.7}{1.4} & \textbf{\mstd{60.7}{1.8}}  & \mstd{58.8}{0.8} & \mstd{56.3}{0.9} \\

& \textbf{50} &
\mstd{75.4}{0.4} & \mstd{75.0}{0.2} & \mstd{71.8}{0.3} & \mstd{67.4}{0.1} &
\mstd{77.2}{0.3} & \textbf{\mstd{77.8}{0.2}} & \mstd{67.6}{0.3} &
\mstd{75.2}{1.2} & \textbf{\mstd{\cellcolor{green!20}{81.0}}{1.6}} & \mstd{75.4}{0.4} & \mstd{75.5}{0.2} \\

\midrule

\multirow{2}{*}{\textbf{ResNet-18}} & \textbf{10} &
\mstd{59.6}{0.2} & \mstd{59.2}{0.18} & \mstd{54.1}{0.6} & \mstd{56.1}{0.5} &
\mstd{55.1}{0.1} & \textbf{\mstd{60.0}{0.6}} & \mstd{53.6}{0.6} &
\mstd{44.9}{0.9} & \textbf{\mstd{57.8}{1.1}} & \mstd{57.0}{1.0} & \mstd{53.3}{1.1} \\

& \textbf{50} &
\mstd{75.5}{0.2} & \mstd{75.5}{0.2} & \mstd{70.9}{0.4} & \mstd{67.2}{0.3} &
\mstd{75.9}{0.5} & \textbf{\mstd{77.2}{0.5}} & \mstd{69.2}{0.9} &
\mstd{73.6}{1.0} & \textbf{\mstd{\cellcolor{green!20}{81.7}}{2.7}} & \mstd{75.3}{0.3} & \mstd{69.0}{0.4} \\

\midrule

\multirow{2}{*}{\textbf{ResNet-50}} & \textbf{10} &
\mstd{35.6}{2.3} & \mstd{46.6}{0.6} & \mstd{42.0}{0.9} & \textbf{\mstd{47.4}{0.6}} &
\mstd{38.2}{2.4} & \mstd{44.2}{1.9} & \mstd{37.3}{0.4} &
\mstd{35.9}{1.8} & \mstd{44.1}{2.5} & \textbf{\mstd{47.5}{1.1}} & \mstd{35.4}{3.2} \\

& \textbf{50} &
\mstd{71.0}{0.1} & \mstd{70.6}{0.5} & \mstd{66.5}{0.4} & \mstd{63.4}{0.1} &
\mstd{70.7}{0.1} & \textbf{\mstd{71.3}{0.4}} & \mstd{63.9}{0.7} &
\mstd{66.3}{1.4} & \textbf{\mstd{\cellcolor{green!20}{75.7}}{1.6}} & \mstd{69.1}{0.9} & \mstd{66.9}{0.5}\\

\midrule

\rowcolor{gray!20}\textbf{Student Model} & \textbf{IPC} & \multicolumn{11}{c}{\textsc{\textbf{\large {Evaluation Protocol: Single teacher soft label (RDED)}}}}\\
\cmidrule{1-13}

\multirow{2}{*}{\textbf{ResNet-18}} & \textbf{10} &
\mstd{59.6}{2.6} & \mstd{59.9}{1.9} & \mstd{56.5}{2.6} & \mstd{54.9}{1.9} &
\mstd{58.2}{0.7} & \mstd{59.6}{3.9} & \textbf{\mstd{63.4}{1.1}} &
\mstd{61.4}{2.1} & \textbf{\mstd{\cellcolor{green!20}{65.5}}{1.7}} & \mstd{60.9}{1.2} & \mstd{61.1}{1.3} \\

& \textbf{50} &
\mstd{82.6}{1.6} & \mstd{77.9}{2.1} & \mstd{81.9}{1.0} & \mstd{78.3}{0.4} &
\textbf{\mstd{83.3}{0.5}} & \mstd{82.4}{0.7} & \mstd{80.1}{0.8} &
\textbf{\mstd{\cellcolor{green!20}{85.8}}{0.3}} & \mstd{85.1}{1.8} & \mstd{85.1}{1.3} & \mstd{84.9}{0.7} \\

\midrule

\multirow{2}{*}{\textbf{ResNet-50}} & \textbf{10} &
\mstd{54.8}{1.7} & \mstd{54.6}{1.5} & \mstd{51.7}{0.8} & \mstd{51.7}{2.2} &
\mstd{54.8}{1.3} & \textbf{\mstd{57.2}{0.6}} & \mstd{56.4}{0.2} &
\mstd{54.8}{0.9} & \textbf{\mstd{\cellcolor{green!20}{63.1}}{5.7}} & \mstd{58.9}{0.5} & \mstd{58.0}{2.4} \\

& \textbf{50} &
\mstd{80.6}{1.1} & \mstd{76.2}{0.7} & \mstd{78.6}{2.4} & \mstd{77.0}{1.9} &
\textbf{\mstd{81.2}{0.9}} & \mstd{79.3}{2.4} & \mstd{79.3}{1.4} &
\mstd{83.0}{0.9} & \textbf{\mstd{\cellcolor{green!20}{84.7}}{1.8}} & \mstd{83.3}{0.9} & \mstd{82.4}{0.9} \\

\midrule

\rowcolor{gray!20}\textbf{Student Model} & \textbf{IPC} & \multicolumn{11}{c}{\textsc{\textbf{\large {Evaluation Protocol: Multiple teachers soft label (EDC)}}}}\\
\cmidrule{1-13}

\multirow{2}{*}{\textbf{ResNet-18}} & \textbf{10} &
\mstd{62.61}{2.29} & \mstd{63.58}{0.97} & \mstd{58.73}{0.81} & \mstd{62.78}{0.20} &
\mstd{62.06}{1.46} & \textbf{\mstd{67.11}{0.52}} & \mstd{65.42}{0.26} &
\mstd{63.56}{0.20} & \mstd{64.44}{1.68} & \textbf{\mstd{65.88}{0.44}} & \mstd{63.87}{2.34} \\

& \textbf{50} &
\mstd{83.52}{0.64} & \mstd{81.18}{0.21} & \mstd{81.00}{0.35} & \mstd{78.46}{0.58} &
\textbf{\mstd{84.64}{0.26}} & \mstd{84.54}{0.44} & \mstd{79.41}{0.34} &
\textbf{\mstd{\cellcolor{green!20}{84.76}}{0.36}} & \mstd{84.73}{0.49} & \mstd{84.70}{0.36} & \mstd{84.37}{0.90} \\

\midrule

\multirow{2}{*}{\textbf{ResNet-50}} & \textbf{10} &
\mstd{59.40}{1.99} & \mstd{60.51}{0.95} & \mstd{56.23}{1.08} & \mstd{58.74}{1.78} &
\mstd{59.18}{1.05} & \textbf{\mstd{63.52}{0.41}} & \mstd{62.76}{0.95} &
\mstd{58.87}{1.17} & \mstd{62.27}{0.83}  & \textbf{\mstd{62.80}{0.94}} & \mstd{61.41}{2.55} \\

& \textbf{50} &
\mstd{82.37}{0.37} & \mstd{80.98}{0.41} & \mstd{79.67}{0.29} & \mstd{77.66}{0.21} &
\mstd{83.86}{0.45} & \textbf{\mstd{84.23}{0.52}} & \mstd{78.74}{0.65} &
\mstd{84.60}{0.11} & \mstd{83.85}{0.68}  & \textbf{\mstd{\cellcolor{green!20}{84.81}}{0.14}} & \mstd{84.38}{0.46} \\

\bottomrule[1.25pt]
\end{tabular}
}
}
% \vspace{-15pt}
\end{table}

\subsection{Efficiency and Quality Analysis of Condensed Sets}
\label{sec:efficiency_and_quality}
\begin{wraptable}{r}{0.3\linewidth}
    \centering
    \vspace{-0.5cm}
\resizebox{\linewidth}{!}{
    \begin{tabular}{rrr}
    \toprule[1.25pt]
     \textbf{\large Method} & \multicolumn{2}{c}{\textbf{\large Time (hrs)}}  \\
     \midrule
        \rowcolor{gray!20} &\textbf{10 IPC}& \textbf{50 IPC}\\
     \cmidrule(ll){2-3}
     \cellcolor{Cerulean!30}ManifoldGD~\cite{roy2026manifoldgd} & 12.8 & 51.7\\
     \cellcolor{Cerulean!30}MGD~\cite{chan2025mgd3} & 11.8 & 50.7 \\
     \cellcolor{Cerulean!30}DiT~\cite{peebles2023scalable} & 1.4 & 7.0\\
     \cellcolor{Cerulean!30}FADRM+~\cite{cui2026fadrm} & 2.5 & 12.5\\
     \cdashline{1-3} 
     % \cmidrule{2-3}
      \Tstrut
     \cellcolor{RubineRed!20} Clustering & 2.0 & 2.0\\
     \cellcolor{RubineRed!20}AUM~\cite{pleiss2020identifying} & 6.0 & 6.0\\
     \cellcolor{RubineRed!20}Concepts~\cite{mehra2025coreset} & 3.0 & 3.0\\
     \bottomrule[1.25pt]
    \end{tabular}
    }
    \caption{Wall clock time for condensed set creation via CS \& DD methods for ImageNet-1K.}
    \label{tab:time-complexity-table}
    \vspace{-15pt}
\end{wraptable}

\textbf{Time Complexity Analysis.} 
% DD approaches exhibit substantial variation in computational cost, largely determined by whether they require a generative model pre-trained on the full training set. 
Training‑based DD methods (e.g., VLCP, D3HR, Minimax) are among the costliest DD methods due to their requirement of training or finetuning of large generative models. 
Training‑free methods reduce this cost but still depend on generative models or incur fixed preprocessing overhead followed by sampling which scales with IPC (e.g., ManifoldGD, MGD, DiT‑Distillation). Decoupled approaches such as FADRM+ avoid generative models but require pretrained classifiers. In contrast, coreset selection relies on lightweight scoring or proxy models. Methods such as Clustering, AUM and Concepts incur a one‑time cost that is effectively independent of IPC, yielding substantially lower and more scalable construction cost in practice.
% Training-based DD methods such as VLCP, D3HR, and Minimax not only depend on a pre-trained generative model but also require additional finetuning during distillation, making them the most time-intensive. 
% Training-free DD methods remove the finetuning stage but still rely on a pre-trained generative model. 
% Purely sampling-based DD methods such as DiT-Distillation result in strictly linear scaling with IPC. 
% Other training-free DD methods such as ManifoldGD and MGD incur a fixed preprocessing overhead (e.g., clustering or manifold construction), followed by a sampling stage that also scales linearly with IPC. 
% Decoupled approaches such as FADRM+ avoid generative models altogether but instead require pre-trained classifiers on the training data, yielding lower and more predictable runtimes. 

% In contrast, CS methods follow fundamentally different time-complexity patterns because they rely on scoring or proxy-based selection rather than synthesis.
% Methods such as AUM incur a one-time cost for training a lightweight proxy model to record training dynamics and computing importance scores; this cost is independent of the target IPC, resulting in effectively constant selection time. 
% Similarly, lightweight CS approaches such as Concepts achieve constant or near-constant selection complexity by training only a small linear model. 
Table~\ref{tab:time-complexity-table} summarizes the synthesis and selection times of DD and CS methods on ImageNet-1K. 
CS methods~\textit{(pink)} exhibit constant runtime across different IPC values, whereas DD methods~\textit{(blue)} scale with IPC. Moreover, ManifoldGD and MGD~\textit{(first two rows)} incur substantial overhead due to expensive pre-processing steps, including clustering and manifold projection, in addition to the DiT sampling cost~\textit{(row 3)}. 
Finally, FADRM+ is efficient at small IPC values as it avoids generative modeling, but its runtime increases with IPC. \underline{Note.} 
In this analysis, we have omitted any time required for training any generative/classification model needed by DD or CS methods that are available off-the-shelf such as DiT-XL/2 needed on training-free DD methods or teacher models used in recover stage of SRe2L-based methods.

{\bf Representativeness and Diversity.}
Along with downstream accuracy, we analyze the quality and structure of condensed datasets using metrics such as representativeness, diversity, and FID, together with qualitative visual inspection. 

\begin{wrapfigure}{l}{0.25\textwidth}
\vspace{-0.35cm}
\centering{\includegraphics[width=.25\textwidth]{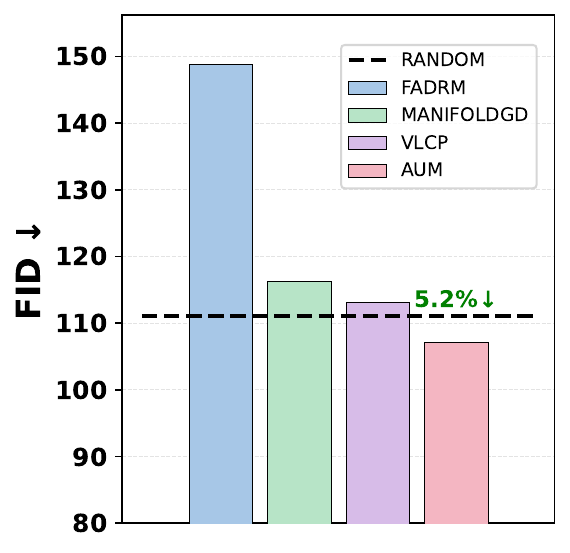}
  \caption{FID comparison for 50 IPC condensed datasets for ImageNet-1K.}
  \label{fig:imagenet1k_fid50}
  }
  \vspace{-15pt}
\end{wrapfigure}
Representativeness and diversity are computed using cosine similarity in a fixed feature space extracted from a ResNet‑18 ($f$) pretrained on ImageNet‑1K similar to \cite{roy2026manifoldgd}. 
Let $\mathcal{T}$ denote the condensed set and $\mathcal{D}_{\text{val}}$ be the validation set. Representativeness is defined as
\(
\min_{j} \max_{i} \; \cos\!\left(f_i^{\mathcal{T}}, f_j^{\mathcal{D}_{\text{val}}}\right),
\)
capturing how well $\mathcal{T}$ covers the original data distribution.
Diversity is defined as
\(
 1 - \max_{i \neq j} \; \cos\!\left(f_i^{\mathcal{T}}, f_j^{\mathcal{T}}\right),
\)
measuring redundancy within the condensed set. Higher representativeness indicates better coverage, while higher diversity reflects a greater sample spread.
We report an average for all the metrics computed class-wise for 50 IPC sets for one representative method from each class of DD and CS methods in Table~\ref{tab:related_methods}.
% , and report the average results. 
Across the 3 datasets, the representativeness–diversity scatter plots (Figs.~\ref{fig:teaser_subfig2},~\ref{fig:ipc50_rep_div}, and~\ref{fig:rep_div_ipc10} in the App.~\ref{app:additional_results}) consistently show that AUM occupies the upper‑right region of the trade‑off space with simultaneously high representativeness and high diversity. 
In contrast, DD methods which are competitive to CS methods in representativeness, lag in diversity, indicating that they concentrate probability mass on a narrower set of visual modes. FADRM+, in particular, exhibits noticeably lower diversity for datasets with many classes, while ManifoldGD and VLCP fall in between, depending on the dataset.

% These quantitative trends align closely with the FID results (Figs.~\ref{}), where AUM consistently attains the lowest FID across datasets, reflecting superior alignment with the real data distribution. 
% DD methods such as VLCP and ManifoldGD often achieve reasonable FID but trail AUM, while FADRM exhibits systematically higher FID, suggesting artifacts or distributional mismatch in the synthesized samples.

{\bf Qualitative Comparison.} 

\begin{wrapfigure}{r}{0.55\textwidth}
\vspace{-0.6cm}
\centering{\includegraphics[trim={1.8cm 0 0 0},clip, width=.5\textwidth]{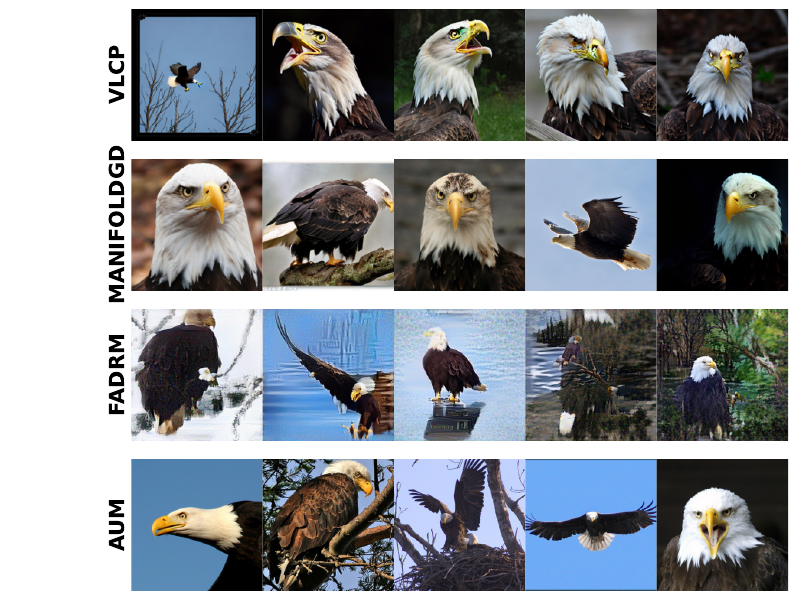}
  \caption{Coreset created by AUM exhibits greater intra‑class diversity, with variations in pose, viewpoint, and background. DD methods, however, repeat similar canonical views or have visible artifacts, hurting their diversity and representativeness.}
  \label{fig:qual_5_eagle}
  }
  \vspace{-10pt}
\end{wrapfigure}
Fig.~\ref{fig:qual_5_eagle} shows representative samples for the \textit{Bald Eagle} class which shows that AUM selects a diverse set of real images, spanning variations in pose, background, viewpoint, and lighting, indicating broader coverage of intra‑class structure. In contrast, VLCP and ManifoldGD frequently repeat highly similar eagle configurations, with the same posture or background recurring multiple times, reflecting reduced diversity. 
FADRM+ additionally exhibits visible artifacts and unnatural textures in several examples. %, consistent with its inferior diversity and FID scores (see Fig.~\ref{fig:fid_ips50}). 
While distilled datasets need not be visually realistic, reduced visual diversity and repeated canonical patterns often reflect limited coverage of the underlying data distribution, which is consistent with the observed diversity and FID trends (see Figs.~\ref{fig:imagenet1k_fid50},~\ref{fig:fid_ipc50_others} in the App.~\ref{app:additional_results}).
These results highlight CS methods produce condensed sets that better preserve the structure and variability of the original data distribution. 

% Qualitative comparison. Fig.~\ref{fig:qual_5_eagle} shows representative samples for the \textit{Bald Eagle}  class. AUM selects a diverse set of real images, capturing variations in pose, viewpoint, background, and illumination, which indicates broader coverage of intra‑class structure. In contrast, VLCP and ManifoldGD frequently repeat visually similar eagle configurations, with the same postures or backgrounds appearing multiple times, reflecting reduced diversity. FADRM further exhibits visible artifacts and unnatural textures in several samples. While distilled datasets are not required to be visually realistic, repeated canonical patterns and reduced visual diversity often signal limited coverage of the underlying data distribution, consistent with the observed diversity and FID trends (see Fig.~\ref{fig:fid_ips50}). Overall, these observations highlight that coreset selection methods produce condensed datasets that better preserve the structure and variability of the original data distribution.

% These observations highlight that perceptual sharpness alone does not guarantee diversity or representational coverage, and that synthetic condensation methods may overfit to a narrow subset of visual modes.

% This pattern aligns with the FID comparisons at IPC = 50, where AUM consistently achieves the lowest FID across ImageNet‑1K, ImageNet‑100, and ImageNette, reflecting better alignment with the original data distribution.

% We measure coverage and spread using the representativeness and diversity metrics adopted in MGD and ManifoldGD. Both metrics 

% \AM{include FID}

\section{Conclusion and Limitations}
\label{sec:conclusion}
% In the context of data‑efficient learning, dataset distillation (DD) has emerged as a principled approach to data compression, aiming to replace large training datasets with compact synthetic samples. However, prior work has evaluated DD methods under inconsistent evaluation protocols, with some studies relying on single‑teacher or multi‑teacher knowledge distillation pipelines, while others use standard empirical risk minimization. This lack of standardized evaluation makes it difficult to disentangle the intrinsic effectiveness of distilled datasets from evaluation‑specific supervision effects. Moreover, many DD methods explicitly claim superiority over coreset selection (CS), arguing that restricting condensed datasets to subsets of real samples fundamentally limits expressiveness, yet such claims are rarely substantiated through controlled empirical comparisons.
In this work, we systematically evaluated the practical effectiveness of DD and the quality of distilled data under controlled and standardized settings. 
We benchmarked SOTA DD methods under three commonly used evaluation protocols and directly compared them against CS approaches. 
Our results show that the performance of DD methods is strongly influenced by the evaluation protocol, with substantial variation in absolute accuracy across hard‑label, single‑teacher, and multi‑teacher settings. 
While distilled datasets can be competitive on small datasets such as ImageNette at low IPC, their advantages diminish on larger datasets such as ImageNet-100 and ImageNet-1K. 
Beyond accuracy, distilled datasets perform poorly in terms of representativeness and diversity and incur significantly higher construction costs, often requiring training large generative models or sampling procedures that scale linearly with IPC. 
In contrast, CS achieves competitive or superior performance while remaining computationally efficient, IPC‑agnostic, and easier to deploy across domains. 
These findings suggest that future progress in DD must be grounded in consistent evaluation across method types in Table~\ref{tab:related_methods} and justified not only by comparison to DD methods, but also by comparison to strong coresets with hard label and at least one soft-label evaluation protocol. 
This is necessary because evaluation‑specific supervision can otherwise distort conclusions about the effectiveness of distilled data, echoing reproducibility concerns raised across ML more broadly.

{\bf Limitations: }
In this work, we focus on image classification, as it is the primary task considered in prior DD works. 
Within classification, our analysis is restricted to ImageNet‑1K and its subsets, since many DD methods release distilled datasets only for this dataset. 
Moreover, since several DD approaches rely on generative models trained on full datasets, it limits their applicability to other domains and makes it impractical to generate distilled sets for these new domains, for our evaluation. 

\section{Acknowledgment}
We thank the anonymous reviewers for their insightful comments and valuable suggestions.
% to obtain distilled datasets for additional tasks.
% In this work, we focus on image classification, as it is the primary task considered in prior DD works. 
% Within classification, our analysis is restricted to ImageNet‑1K and its subsets, since many DD methods released subsets for this dataset. Moreover since the some DD methods require generative models pretrained on ImageNet‑1K, which makes it difficult to use these DD methods to obtain distilled datasets for our evaluation for other domains. 
% This choice was made to enable direct and fair comparison across distilled datasets released by prior works.
% under consistent experimental settings.
% The extent to which our findings generalize to other datasets or tasks remains an open question.

\bibliographystyle{unsrt}
\bibliography{main}

%%%%%%%%%%%%%%%%%%%%%%%%%%%%%%%%%%%%%%%%%%%%%%%%%%%%%%%%%%%%

\appendix

\clearpage
\appendix
\onecolumn
\begin{center}
{\LARGE \bf Appendix}
\end{center}

We present additional details of DD and CS methods, followed by details of the algorithm used for sampling the coreset, and additional experiments which were omitted in the main paper due to space restrictions. 
% \AM{Add coresets of concepts and AUM to supplementary}

\section{Additional related work and details of DD and CS methods}
\label{app:additional_rw}

{\bf DD methods.}
Recently many DD methods have been proposed that use generative models. These approaches leverage powerful generative models, such as GANs and diffusion models, to synthesize distilled datasets. 
Early works in this direction, such as gradient matching-based dataset condensation~\cite{zhao2021dataset} and trajectory matching methods such as MTT~\cite{cazenavette2022dataset}, directly optimized synthetic data to mimic training dynamics of real data. 
Subsequent works, explored feature-space alignment~\cite{zhao2023dataset} and generative adversarial formulations such as GAN-IT~\cite{zhao2022synthesizinginformativetrainingsamples}, improving scalability and sample fidelity.
More recent advances have been driven by diffusion models.

{\bf CS methods.}
A popular line of work in CS leverages {\em geometric or diversity-based criteria}, such as k-center greedy selection~\cite{sener2017active} or clustering-based methods~\cite{feldman2020turning}, to ensure coverage of the data distribution. 
Although computationally more efficient than training-dynamics-based approaches discussed in the main paper, these methods often rely on pairwise distance computations, which can become prohibitive at scale and may not capture semantic importance of samples.  
More recently, {\em gradient- and influence-based methods}~\cite{koh2017understanding,killamsetty2021gradmatch} have been proposed that estimate the contribution of each sample to model training by approximating its effect on the loss or gradients. While theoretically grounded, these methods typically involve higher-order computations or repeated optimization, limiting their scalability.  
Another line of works in CS focus on the problem of \textit{adaptive subset selection} \cite{killamsetty2023milo, tukan2023provable, killamsetty2021grad}. 
Unlike CS, these works improve the training convergence of a model by selecting a new subset of data for every epoch of model training. 
% Thus, unlike our work, these works do not prune the dataset but rather keep selecting small, potentially non-overlapping, subsets every few epochs.
% Works in this area include \cite{killamsetty2023milo, tukan2023provable}, which select new subsets by using  downstream model unlike some other works such as GradMatch \cite{killamsetty2021grad}, which select subsets based on the downstream model. 
% While our work does not target adaptive subset selection, we evaluated how well our approach performs without any change on this problem in App.~\ref{app:adaptive_cs}. Our results show that, even on for this application, our concept-based score is an effective method. %yields performance comparable to existing approaches.

{\bf Details of the concept-based method for CS.}
The approach proposed in \cite{mehra2025coreset} leverages pretrained models to evaluate the importance of each training sample for CS. 
Specifically, for an image $x$, it works by extracting visual embeddings $\mathcal{V}_{enc}(x) \in \mathbb{R}^d$ using a pretrained multimodal encoder (e.g., CLIP) and a matrix of $N_C$ concept embeddings mapped into the same embedding space, generated via an LLM denoted by $E_C \in \mathbb{R}^{N_C \times d}$, then the method computes a concept similarity vector for $x$ as $g(x; E_C) = \mathcal{V}_{enc}(x) E_C^\top$.
% denote a matrix of $N_C$ concept embeddings generated using an LLM and mapped into the same embedding space. 
% The concept similarity vector for $x$ is computed as $g(x; E_C) = \mathcal{V}_{enc}(x) E_C^\top$.
Then it obtains class-wise predictions by training a linear concept bottleneck classifier
$f(x; W) = g(x; E_C) W^\top$,
where $W \in \mathbb{R}^{|\mathcal{Y}| \times N_C}$ using the cross-entropy loss over the training set for $T$ epochs, while keeping the encoders and concept embeddings fixed.
During training of this linear layer the margin of a sample $(x,y)$ at epoch $t$ denoted as $M_{\text{concept}}^t(x,y) = f_y^t(x) - \max_{y' \neq y} f_{y'}^t(x)$ is recorded. 
Finally, the \emph{concept-guided AUM score} after $T$ epochs of training is computed as
% \begin{equation}
\(\mathrm{AUM}_{\text{concept}}(x,y) = \frac{1}{T} \sum_{t=1}^T M_{\text{concept}}^t(x,y).\)

\section{Algorithm for stratified sampling for CS}
\label{app:algo_ccs}
A crucial step in CS is the sampling strategy used to form the coresets.
This becomes especially important at extremely small data budgets, where how the samples are chosen to form the coreset across the data distribution is as important as the scoring function. 
A naïve top‑k selection could over‑represent either outliers or highly redundant examples, leading to poor coverage and reduced generalization. 
Consequently, modern CS methods emphasize coverage‑centric sampling strategies that balance representativeness and diversity across difficulty levels.

Alg.~\ref{alg:ipc_ccs} adapts the coverage‑centric coreset selection (CCS) \cite{zheng2022coverage} paradigm to the images‑per‑class (IPC) setting required for class‑balanced condensation. 
Specifically, we extend CCS to operate independently per class and enforce an explicit IPC budget. A critical component in CCS is the misratio (cutoff ratio $\beta$), which removes overly hard or noisy samples prior to stratified sampling, preventing the IPC budget from being wasted on mislabeled or uninformative outliers. 
This misratio mechanism is central to both the original CCS formulation and prior concept‑based method. 
In the following section, we explicitly ablate the misratio parameter to quantify its impact on performance.

\subsection{Ablation on impact of misratio \texorpdfstring{($\beta$)}{(beta)}}

We study the effect of the hard cutoff rate used during sampling when forming coresets. 
Table~\ref{tab:misratio_ablation} reports results across ImageNette, ImageNet‑100, and ImageNet‑1K for IPC values of 10 and 50. 
Across all datasets and two CS methods (AUM and Concepts), we observe in the low‑IPC regime, when the coreset budget is small (e.g., IPC=10), higher misratios (such as 0.9), which remove a larger fraction of the hardest samples, tend to yield better performance. 
This suggests that, under severe data scarcity, retaining comparatively easier and more reliable samples leads to improved coverage of the underlying data distribution and more stable downstream training. 
% In contrast, aggressively retaining hard samples in this regime appears to amplify noise and coverage gaps, resulting in degraded accuracy.
As IPC increases (e.g., IPC=50), the performance peaks at moderate cutoff values (such as 0.5 and 0.7). 
This indicates that with a larger data budget, the coreset benefits from including more challenging samples improving distributional coverage. 
These observations are consistent with prior findings on coverage‑centric coreset selection \cite{zheng2022coverage,mehra2025coreset}, which show that optimal hard cutoff rates tend to increase as the pruning budget becomes more constrained. 
% Intuitively, when the total data budget is limited, shifting more sampling capacity toward high‑density strata by filtering out unreliable hard samples—improves coverage and downstream performance.
% Overall, this ablation highlights that the optimal misratio depends on the target IPC, with smaller subsets benefiting from more conservative (higher) cutoff rates. This further emphasizes the importance of coupling importance scoring with coverage‑aware sampling, especially in the low‑data regime that is central to dataset distillation and data‑efficient learning.

We used the optimal misratio identified using this analysis (using the hard-label protocol i.e. \emph{no student–teacher} evaluation setting) for AUM and Concepts and used them for creating the coresets used for various evaluations in the paper. 
For the clustering-based method we used the misratio of $0.9$ for IPC=10 and $0.5$ for IPC=50 for all datasets.

\section{Additional experimental results}
\label{app:additional_results}

\begin{algorithm}[t]
\small
\caption{IPC-based Coverage-centric Coreset Selection (IPC-CCS)}
\label{alg:ipc_ccs}
\textbf{Input}: 
Per-class dataset with difficulty scores
$\mathbb{D}_c = \{(x_i, y_i, s_i)\}_{i=1}^{n_c}$,
images per class (IPC) $k$,
cutoff ratio $\beta \in (0,1)$,
number of bins $b$. \\
\textbf{Output}:
Coreset for class $c$: $\mathcal{S}_c$
\begin{algorithmic}[1]
\STATE \# Prune the hardest samples
\STATE Sort $\mathbb{D}_c$ in ascending order of difficulty score $s$
\STATE Remove the lowest $\lfloor \beta \times n_c \rfloor$ samples
\STATE Denote the remaining dataset as $\mathbb{D}_c'$

\STATE \# Stratify remaining samples by difficulty
\STATE Partition the scores in $\mathbb{D}_c'$ into $b$ disjoint bins
\STATE $\mathcal{B} \leftarrow \{B_1, B_2, \dots, B_b\}$, where each $B_i$ contains samples with scores in bin $i$

\STATE Initialize $\mathcal{S}_c \leftarrow \emptyset$
\STATE Initialize $\; m \leftarrow k$
\WHILE{$\mathcal{B} \neq \emptyset$ \AND $m > 0$}
    \STATE Select the bin with the fewest samples:
    \[
    B_{\min} \leftarrow \arg\min_{B \in \mathcal{B}} |B|
    \]
    \STATE Allocate the sampling budget for this bin:
    \[
    m_B \leftarrow \min\left(|B_{\min}|,\; \left\lfloor \frac{m}{|\mathcal{B}|} \right\rfloor\right)
    \]
    \STATE Uniformly sample $m_B$ samples from $B_{\min}$ and add them to $\mathcal{S}_c$
    \STATE $\mathcal{B} \leftarrow \mathcal{B} \setminus \{B_{\min}\}$
    \STATE $m \leftarrow m - m_B$
\ENDWHILE
\STATE \textbf{return} $\mathcal{S}_c$
\end{algorithmic}
\end{algorithm}

\begin{table*}[t]
\centering
\caption{\textbf{Misratio Ablation for Sampling Coresets.} Accuracy (\%) for five misratio values~(0.1, 0.3, 0.5, 0.7 and 0.9) for all three datasets evaluated with the hard-label protocol with ResNet-18 as the student model.}
\label{tab:misratio_ablation}
{\renewcommand{\arraystretch}{1}
\resizebox{0.99\textwidth}{!}{
\begin{tabular}{rcccccc@{\hspace{1.2cm}}ccccc@{\hspace{1.2cm}}ccccc}
\toprule[1.25pt]
% \rowcolor{gray!60} & & \multicolumn{15}{c}{\large \textbf{Datasets and Misratios}} \\
% \cmidrule(lr){2-16}
% \midrule

\rowcolor{gray!20} & & 
\multicolumn{5}{c}{\textbf{ImageNette}} &
\multicolumn{5}{c}{\textbf{ImageNet-100}} &
\multicolumn{5}{c}{\textbf{ImageNet-1K}} \\
\cmidrule{3-7}\cmidrule{8-12}\cmidrule{13-17}

\rowcolor{gray!20}\textbf{Method} & \textbf{IPC} &
0.1 & 0.3 & 0.5 & 0.7 & 0.9 &
0.1 & 0.3 & 0.5 & 0.7 & 0.9 &
0.1 & 0.3 & 0.5 & 0.7 & 0.9 \\
\midrule

\multirow{2}{*}{\textbf{AUM}~\cite{pleiss2020identifying}} & \textbf{10} &
54.3 & 50.9 & \textbf{55.2} & 57.2 & 54.5 &
19.2 & 21.9 & 25.0 & 25.3 & \textbf{27.2} &
10.1 & 12.0 & 15.1 & 17.0 & \textbf{18.4} \\

 & \textbf{50} &
73.6 & \textbf{75.6} & 75.2 & 73.7 & 68.7 &
45.3 & 48.2 & 50.8 & \textbf{51.3} & 49.0 &
40.9 & 42.6 & \textbf{43.5} & 43.0 & 40.4 \\
\midrule

\multirow{2}{*}{\textbf{Concepts}~\cite{mehra2025coreset}} &  \textbf{10} &
51.6 & 51.5 & 49.8 & 51.1 & \textbf{53.6} &
18.4 & 18.2 & 20.4 & \textbf{21.5} & 20.8 &
9.9 & 11.4 & 13.7 & 14.1 & \textbf{14.5} \\

 &  \textbf{50} &
72.9 & 74.3 & 74.2 & 74.3 & \textbf{75.7} &
45.4 & 48.6 & 49.3 & \textbf{50.3} & 50.1 &
40.8 & 42.4 & \textbf{42.7} & 42.4 & 41.0 \\

\bottomrule[1.25pt]
\end{tabular}
}
}
\end{table*}

\subsection{Generalization across student architectures} 
\label{app:architecture_generalization}
\begin{table*}[t] 
\centering 
\caption{ Top-1 accuracy (\%, mean$\pm$std) on ImageNet-1K using Swin and ViT-B-16 student architectures under the RDED evaluation pipeline. 
All student models are trained for 300 epochs using soft labels generated by a ResNet-18 teacher.
The best result in each row is shown in bold. 
} 
\label{tab:architecture_generalization} 
{\renewcommand{\arraystretch}{1.25} 
\resizebox{0.90\textwidth}{!}{ 
\begin{tabular}{rrccc@{\hspace{1.2cm}}c@{\hspace{0.25cm}}c} 
\toprule[1.25pt] & & \multicolumn{3}{c}{ \cellcolor{Cerulean!30}{\textbf{\large Distillation Methods}} } & \multicolumn{2}{c}{ \cellcolor{RubineRed!20}{\textbf{\large Coreset Selection Methods}} } \\ 
\cmidrule(lr){3-5} \cmidrule(lr){6-7} \textbf{Student Model} & \textbf{IPC} & \textbf{VLCP}~\cite{zou2025vlcp} & \textbf{ManifoldGD}~\cite{roy2026manifoldgd} & \textbf{FADRM+}~\cite{cui2026fadrm} & \textbf{Random} & \textbf{AUM}~\cite{pleiss2020identifying} \\ 
\midrule 
\multirow{2}{*}{\textbf{Swin}} & \textbf{10} & \mstd{22.9}{0.6} & \textbf{\mstd{26.2}{0.7}} & \mstd{19.8}{0.3} & \mstd{23.9}{0.3} & \textbf{\mstd{\cellcolor{green!20}{27.2}}{0.6}} \\ 
& \textbf{50} & \mstd{62.2}{0.2} & \textbf{\mstd{63.7}{0.1}} & \mstd{60.6}{0.1} & \mstd{63.7}{0.2} & \textbf{\mstd{\cellcolor{green!20}{64.0}}{0.2}} \\ 
\midrule 
\multirow{2}{*}{\textbf{ViT-B/16}} & \textbf{10} & \mstd{17.8}{0.1} & \textbf{\mstd{20.1}{0.8}} & \mstd{16.4}{0.6} & \mstd{20.5}{0.2} & \textbf{\mstd{\cellcolor{green!20}{23.0}}{0.9}} \\ 
& \textbf{50} & \mstd{61.2}{0.2} & \textbf{\mstd{61.4}{0.2}} & \mstd{58.0}{0.4} & \mstd{62.6}{0.1} & \textbf{\mstd{\cellcolor{green!20}{63.1}}{0.1}} \\ 
\bottomrule[1.25pt] 
\end{tabular} } } 
\end{table*}

%\textbf{\mstd{\cellcolor{green!20}{84.81}}{0.14}}
Here, we extend the evaluation of CS and DD methods to two architectures: ViT-B/16 and Swin. 
We evaluate all methods on ImageNet-1K at 10 and 50 images per class (IPC) using the RDED evaluation protocol. 
Consistent with the main paper, a ResNet-18 teacher provides the soft labels used to train each student architecture. 

Table~\ref{tab:architecture_generalization} reports the resulting top-1 accuracy. 
% The conclusions remain consistent across both architectures and IPC settings. 
AUM achieves the highest accuracy in all four evaluation settings, outperforming the considered dataset distillation methods on both ViT-B/16 and Swin. 
Random selection also remains competitive, outperforming VLCP and FADRM+ in every setting and matches/exceeds the performance of ManifoldGD at 50 IPC. 
These results demonstrate that the competitiveness of computationally efficient CS methods extends to ViT-B/16 and Swin models.

\subsection{Evaluation in very low IPC setting} 
\label{app:low_ipc}
\begin{table*}[t]
\centering
\caption{Top-1 accuracy (\%, mean$\pm$std) on ImageNet-1K at 1 and 5 IPC with RDED. Student models are trained for 300 epochs using soft labels generated by a ResNet-18 teacher. The best result in each row is shown in bold.}
\label{tab:low_ipc_rded}
{\renewcommand{\arraystretch}{1.25}
\resizebox{0.90\textwidth}{!}{
\begin{tabular}{rrccc@{\hspace{1.2cm}}c@{\hspace{0.25cm}}c}
\toprule[1.25pt]
& & \multicolumn{3}{c}{\cellcolor{Cerulean!30}{\textbf{\large Distillation Methods}}} & \multicolumn{2}{c}{\cellcolor{RubineRed!20}{\textbf{\large Coreset Selection Methods}}} \\
\cmidrule(lr){3-5}\cmidrule(lr){6-7}
\textbf{Student Model} & \textbf{IPC} & \textbf{VLCP}~\cite{zou2025vlcp} & \textbf{ManifoldGD}~\cite{roy2026manifoldgd} & \textbf{FADRM+}~\cite{cui2026fadrm} & \textbf{Random} & \textbf{AUM}~\cite{pleiss2020identifying} \\
\midrule
\multirow{2}{*}{\textbf{ResNet-18}} & \textbf{1} & \mstd{5.0}{0.1} & \textbf{\mstd{6.1}{0.6}} & \mstd{5.9}{0.1} & \mstd{5.3}{0.1} & \textbf{\mstd{\cellcolor{green!20}{6.4}}{0.3}} \\
& \textbf{5} & \mstd{28.1}{0.8} & \mstd{31.4}{0.4} & \textbf{\mstd{34.2}{0.2}} & \mstd{29.2}{0.5} & \textbf{\mstd{32.1}{0.6}} \\
\midrule
\multirow{2}{*}{\textbf{ResNet-50}} & \textbf{1} & \mstd{4.7}{0.2} & \mstd{5.3}{0.6} & \mstd{5.4}{0.3} & \mstd{4.6}{0.4} & \textbf{\mstd{\cellcolor{green!20}{6.2}}{0.5}} \\
& \textbf{5} & \mstd{31.6}{0.5} & \mstd{34.4}{1.0} & \textbf{\mstd{37.5}{0.9}} & \mstd{32.0}{2.0} & \textbf{\mstd{35.6}{0.6}} \\
\bottomrule[1.25pt]
\end{tabular}
}
}
\end{table*}
% \AM{Add details for subset construction} 
To evaluate the performance of DD and CS methods for very low IPCs, we evaluate three distillation methods (FADRM+, ManifoldGD, VLCP) and two coreset methods (Random, AUM) for IPC=1 and IPC=5 on ImageNet-1K with ResNet-18 as the teacher model for the RDED (single-teacher soft-label) protocol. 
For this experiment, the condensed sets were created by randomly sampling 1 or 5 images per class from the 50 IPC condensed sets (for AUM we sampled from the 50 IPC coreset created with a misratio of 0.9). 
Our results in Table~\ref{tab:low_ipc_rded} show that the performance of some DD methods does not surpass the performance of random subsets similar to our observation in Table~\ref{tab:imagenet1k-results}. While FADRM+ achieves the best result for IPC 5, CS methods such as AUM remain competitive even in the low IPC regime. 
% similar trend compared to IPC 10 and IPC 50 where we see that computationally cheaper and faster coreset baselines (AUM, Random) outperform the computationally expensive and slower state-of-the-art distillation methods (VLCP, ManifoldGD, FADRM). 

\subsection{Additional results for diversity and representativeness}
\begin{figure}
% \begin{wrapfigure}{r}{0.6\linewidth}
    \centering
    \begin{subfigure}[t]{0.325\linewidth}
        \centering
        \includegraphics[width=\linewidth]{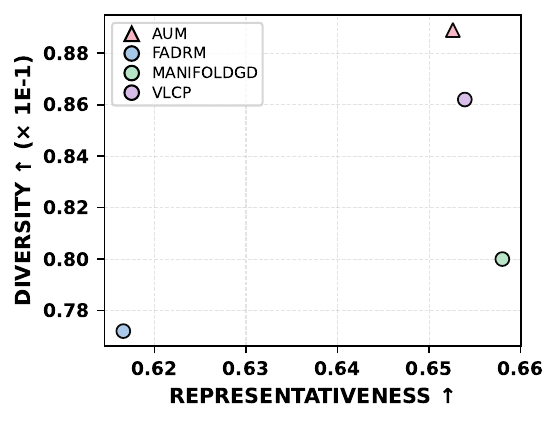}
        \caption{ImageNet-100}
        \label{fig:imagenet100_ips50}
        % \vspace{-10pt}
    \end{subfigure}
    % \hfill
    % \begin{subfigure}[t]{0.24\linewidth}
    %     \centering
    %     % \includegraphics[width=\linewidth]{images/imagenet1k_div_vs_rep_classwise_ipc10.pdf}
    %     \includegraphics[width=\linewidth]{images/imagenet100_fid_barplot_ipc50.pdf}
    %     \caption{}
    %     \label{fig:imagenet100_barplot}
    % \end{subfigure}
    \begin{subfigure}[t]{0.325\linewidth}
        \centering
        \includegraphics[width=\linewidth]{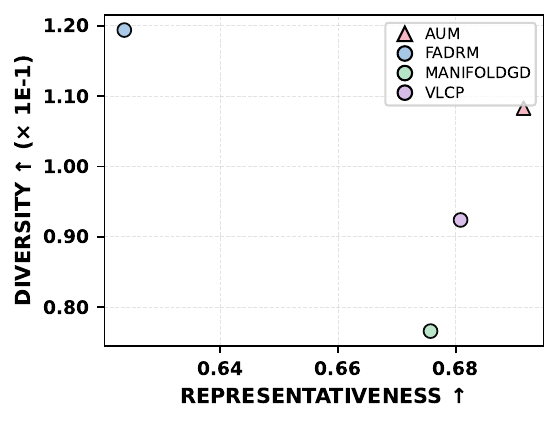}
        \caption{ImageNette}
        \label{fig:imagenette_ips50}
        % \vspace{-10pt}
    \end{subfigure}
    \caption{Representativeness–diversity trade‑off at {\bf 50} IPC for ImageNet‑100, and ImageNette. Results show that AUM achieves a more favorable balance, occupying the upper right corner, compared to SOTA DD methods.}
    \label{fig:ipc50_rep_div}
    % \vspace{-20pt}
% \end{wrapfigure}
\end{figure}

\begin{figure}[t]
    \centering
    \begin{subfigure}[t]{0.325\linewidth}
        \centering
        \includegraphics[width=\linewidth]{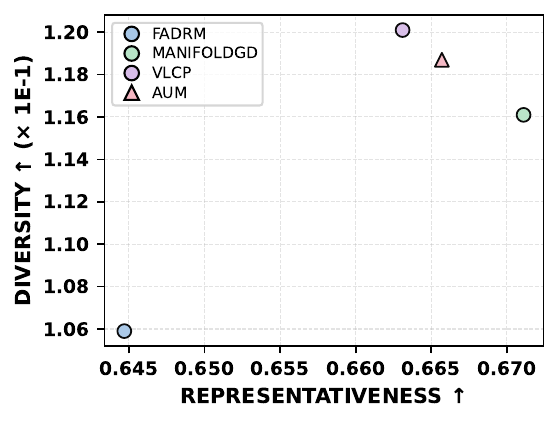}
        \caption{ImageNet-1K}
        \label{fig:imagenet1k_ips10}
    \end{subfigure}
    \hfill
    \begin{subfigure}[t]{0.325\linewidth}
        \centering
        \includegraphics[width=\linewidth]{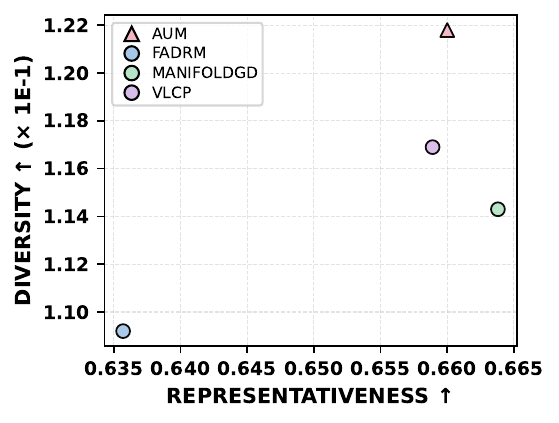}
        \caption{ImageNet-100}
        \label{fig:imagenet100_ips10}
    \end{subfigure}
    \hfill
    \begin{subfigure}[t]{0.325\linewidth}
        \centering
        \includegraphics[width=\linewidth]{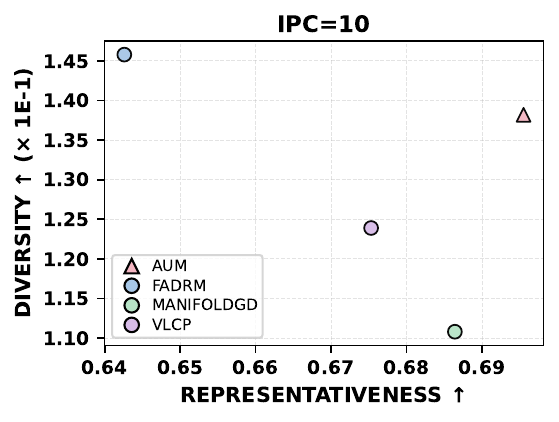}
        \caption{ImageNette}
        \label{fig:imagenette_ips10}
    \end{subfigure}
    \caption{Representativeness–diversity trade‑off at {\bf 10} IPC on ImageNet‑1K, ImageNet‑100, and ImageNette. Similar to the 50 IPC results presented, AUM achieves a more favorable balance between representativeness and diversity, while DD methods exhibit reduced diversity.}
    \label{fig:rep_div_ipc10}
\end{figure}

\begin{table*}[t] 
\centering 
\caption{Representativeness and diversity at 50 IPC across different feature spaces. Higher values are better for both metrics. The best result within each feature space and metric is shown in bold.} 
\label{tab:feature_space_sensitivity} 
\resizebox{\textwidth}{!}{ 
\begin{tabular}{lcccccccccc} 
\toprule & \multicolumn{2}{c}{ResNet-18} & \multicolumn{2}{c}{CLIP ViT-B/32} & \multicolumn{2}{c}{CLIP ViT-L/14} & \multicolumn{2}{c}{SwAV} & \multicolumn{2}{c}{DINO} \\ \cmidrule(lr){2-3} \cmidrule(lr){4-5} \cmidrule(lr){6-7} \cmidrule(lr){8-9} \cmidrule(lr){10-11} Method & Rep. & Div. & Rep. & Div. & Rep. & Div. & Rep. & Div. & Rep. & Div. \\ 
\midrule 
ManifoldGD & 0.6513 & 0.0795 & \textbf{0.6763} & 0.0510 & 0.6743 & 0.0484 & 0.6033 & 0.0903 & 0.3142 & 0.0952 \\ 
VLCP & 0.6514 & 0.0838 & 0.6638 & 0.0525 & 0.6689 & 0.0495 & 0.5916 & 0.0963 & 0.3148 & 0.1030 \\ 
FADRM & 0.6182 & 0.0783 & 0.6318 & 0.0644 & 0.6135 & 0.0459 & 0.5611 & 0.0933 & 0.2526 & 0.1213 \\ 
AUM & \textbf{0.6615} & \textbf{0.0986} & 0.6722 & \textbf{0.0684} & \textbf{0.6776} & \textbf{0.0577} & \textbf{0.6049} & \textbf{0.1043} & \textbf{0.3235} & \textbf{0.1216} \\
\bottomrule 
\end{tabular} } 
\end{table*}
Fig.~\ref{fig:rep_div_ipc10} shows class‑wise representativeness and diversity metrics at 10 IPC for ImageNet‑1K, ImageNet‑100, and ImageNette, complementing the 50 IPC analysis in the main paper. 
Consistent with the trends observed at 50 IPC, AUM maintains comparatively higher diversity while preserving representativeness, despite the extreme sparsity of the condensed datasets.

We further evaluate the representativeness and diversity of the condensed sets for ImageNet-1K using multiple feature spaces beyond the ResNet-18 used in Sec.~\ref{sec:efficiency_and_quality}, including CLIP ViT-B / 32, CLIP ViT-L / 14, SwAV and DINO and present results in Table~\ref{tab:feature_space_sensitivity}. 
The overall trend remains consistent across representations: AUM/coreset selection remains among the strongest methods in both representativeness and diversity, while DD methods generally exhibit lower diversity and/or weaker coverage of the original data distribution.

More specifically, AUM achieves the best or near-best representativeness in ResNet-18, CLIP ViT-L/14, and SwAV, while also achieving the highest diversity in all feature spaces. FADRM consistently obtains the lowest representativeness across feature spaces, while VLCP and ManifoldGD improve representativeness relative to FADRM but remain less diverse than AUM. This is in line with our results in Sec.~\ref{sec:efficiency_and_quality}.

% Overall, these results support the conclusions of Sec.~\ref{sec:efficiency_and_quality} and suggest that our conclusions remain stable across both supervised and self-supervised feature spaces.

\subsection{Comparison of FID for various condensed sets}
\begin{figure}[t!]
    % \centering
    % \begin{subfigure}[t]{0.325\linewidth}
    %     \centering
    %     \includegraphics[width=\linewidth]{images/imagenet1k_fid_barplot_ipc50.pdf}
    %     \caption{ImageNet-1K}
    %     \label{fig:imagenet1k_fid50}
    % \end{subfigure}
    % \hfill
    \centering
    \begin{subfigure}[t]{0.325\linewidth}
        \centering
        \includegraphics[width=\linewidth]{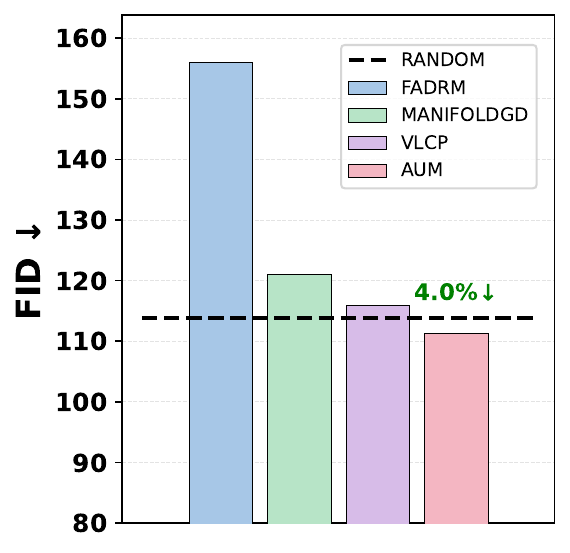}
        \caption{ImageNet-100}
        \label{fig:imagenet100_fid50}
    \end{subfigure}
    % \hfill
    \begin{subfigure}[t]{0.325\linewidth}
        \centering
        \includegraphics[width=\linewidth]{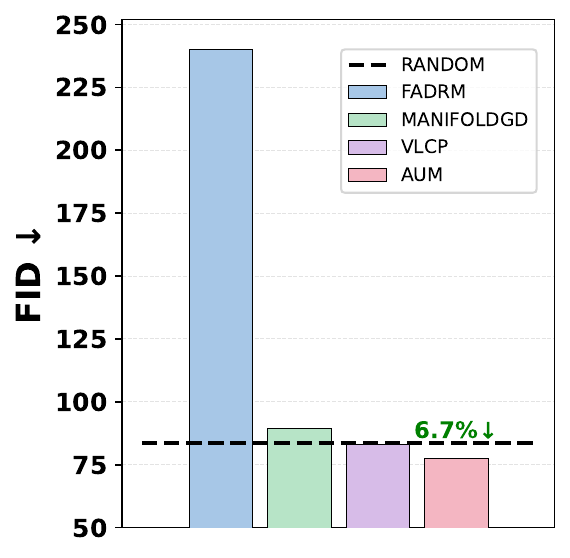}
        \caption{ImageNette}
        \label{fig:imagenette_fid50}
    \end{subfigure}
    \caption{Fréchet Inception Distance (FID) comparison for condensed datasets for {\bf 50} IPC. AUM achieves lower FID than all dataset distillation methods on both datasets and performs better than a randomly chosen subset of data of the same size. This shows that AUM enables finding a subset, highly aligned with the original data distribution. In contrast, several DD methods yield substantially higher FID, suggesting reduced coverage and increased distributional mismatch in the condensed sets.}
    \label{fig:fid_ipc50_others}
\end{figure}

Fig.~\ref{fig:fid_ipc50_others} presents FID results at 50 IPC for ImageNette, ImageNet‑100, and ImageNet‑1K. 
Similar to the findings reported in the main paper, CS methods specifically AUM consistently achieve the lowest FID across all three datasets, indicating superior alignment with the real data distribution.

On ImageNette, AUM improves FID by approximately 6.7\% relative to the best DD method, reflecting better distributional coverage despite the dataset’s small scale and high separability. 
% On ImageNet‑100, the improvement becomes more pronounced, with AUM reducing FID by roughly 4.0\% relative to DD methods.
A similar pattern is observed on ImageNet-100 and ImageNet‑1K, where AUM again achieves the lowest FID, while distillation methods are consistently worse.
These results show that while DD methods can produce visually sharp samples, they can still have distributional mismatch and reduced diversity, particularly on larger and more complex datasets. 

\subsection{Qualitative evaluation of condensed sets}

\begin{figure}[pb!]
    \centering
    \begin{subfigure}[t]{\linewidth}
        \centering
        \includegraphics[width=.85\linewidth]{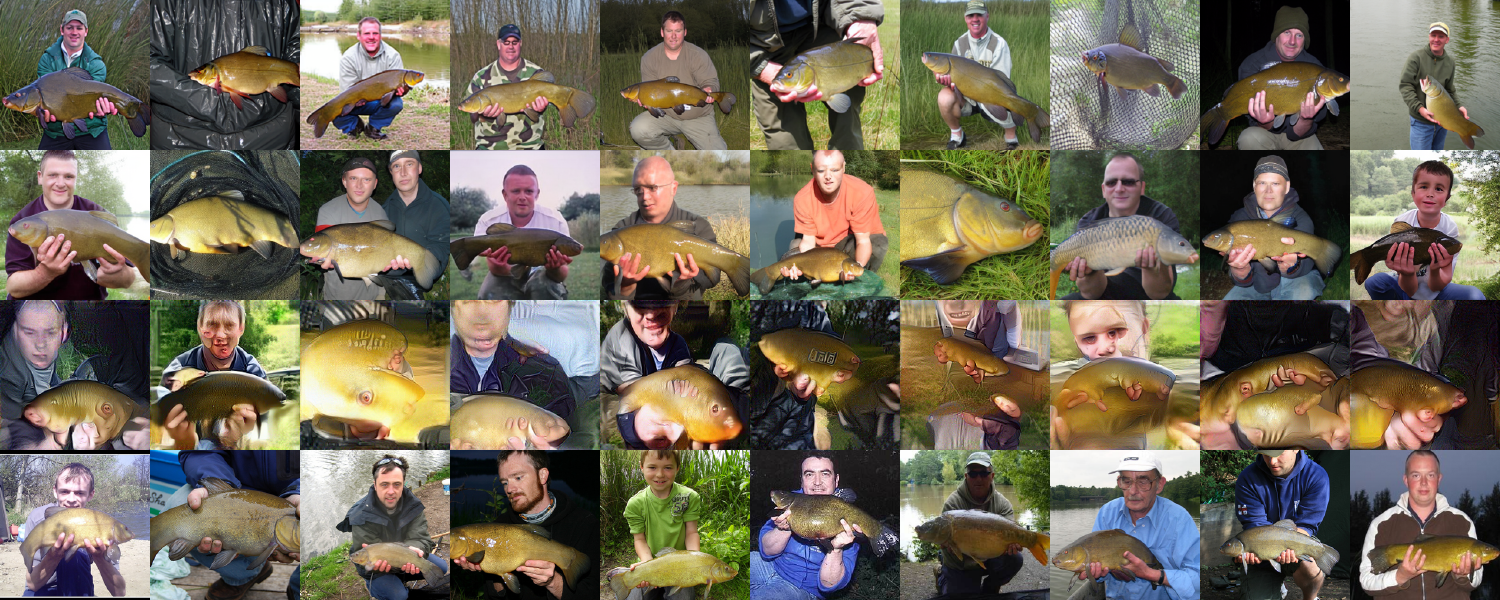}
        \caption{n01440764}
        \label{fig:person_with_fish}
    \end{subfigure}
    \hfill
    \begin{subfigure}[t]{\linewidth}
        \centering
        \includegraphics[width=.85\linewidth]{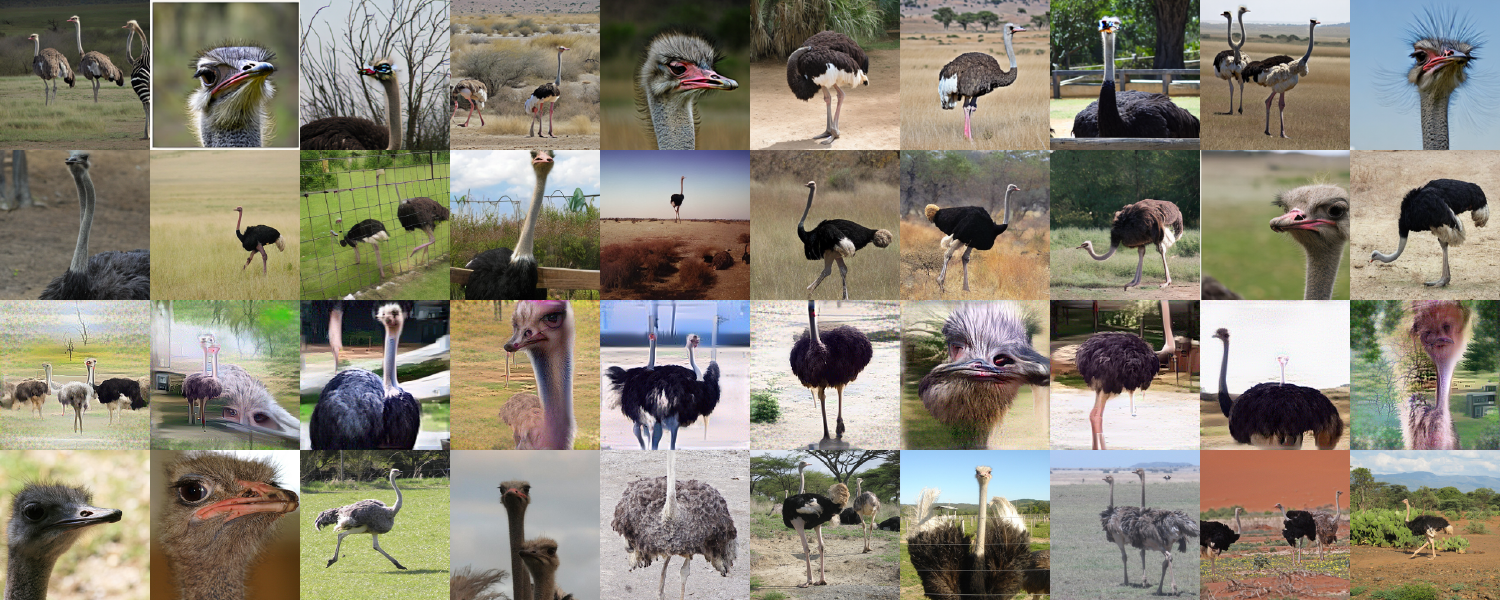}
        \caption{n01518878}
        \label{fig:ostrich}
    \end{subfigure}
    \hfill
    \begin{subfigure}[t]{\linewidth}
        \centering
        \includegraphics[width=.85\linewidth]{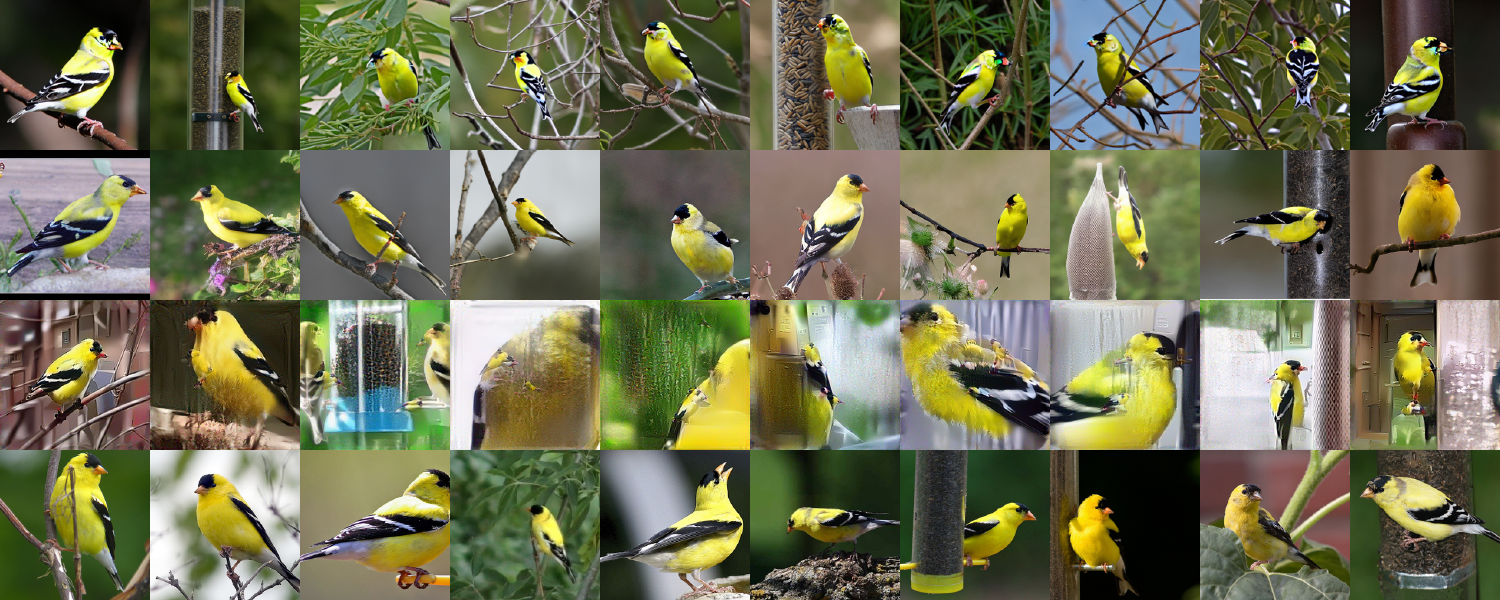}
        \caption{n01531178}
        \label{fig:yellow_bird}
    \end{subfigure}
        \hfill
    \begin{subfigure}[t]{\linewidth}
        \centering
        \includegraphics[width=.85\linewidth]{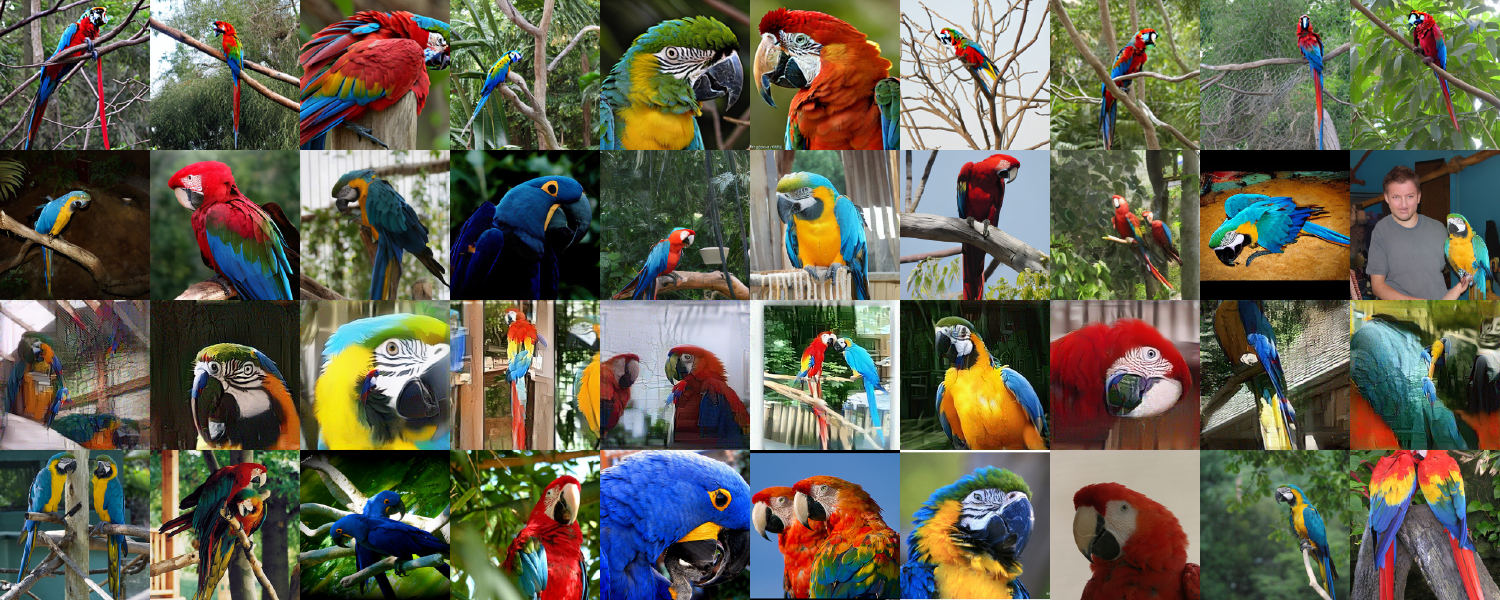}
        \caption{n01818515}
        \label{fig:parrot}
    \end{subfigure}
\caption{10 images per class sampled from condensed datasets produced by VLCP, ManifoldGD, FADRM+, and AUM for 4 synsets of ImageNet-1K.}
\label{fig:grid_pg1}
\end{figure}

\begin{figure}[pb!]
    \centering
    % [pt!]\ContinuedFloat
            \hfill
        \begin{subfigure}[t]{\linewidth}
        \centering
        \includegraphics[width=.85\linewidth]{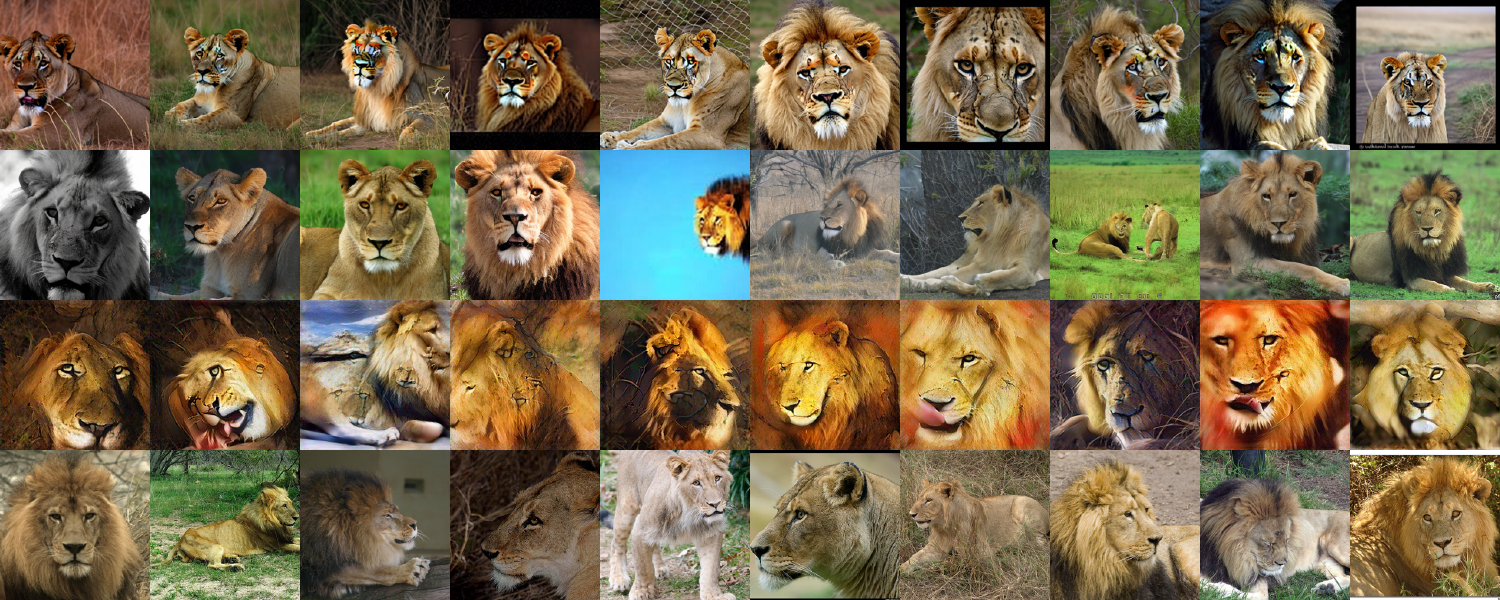}
        \caption{n02129165}
        \label{fig:lion}
    \end{subfigure}
    \hfill
    \begin{subfigure}[t]{\linewidth}
        \centering
        \includegraphics[width=.85\linewidth]{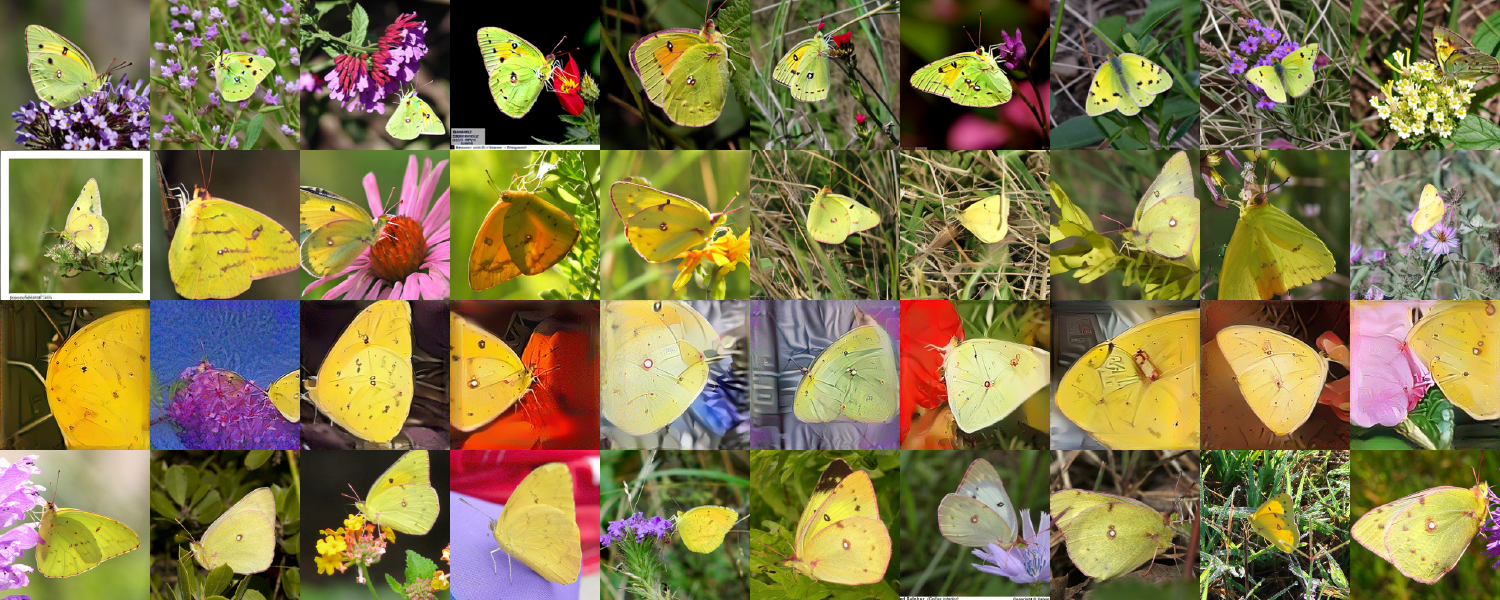}
        \caption{n02281406}
        \label{fig:butterfly}
    \end{subfigure}
    \hfill
    \begin{subfigure}[t]{\linewidth}
        \centering
        \includegraphics[width=.85\linewidth]{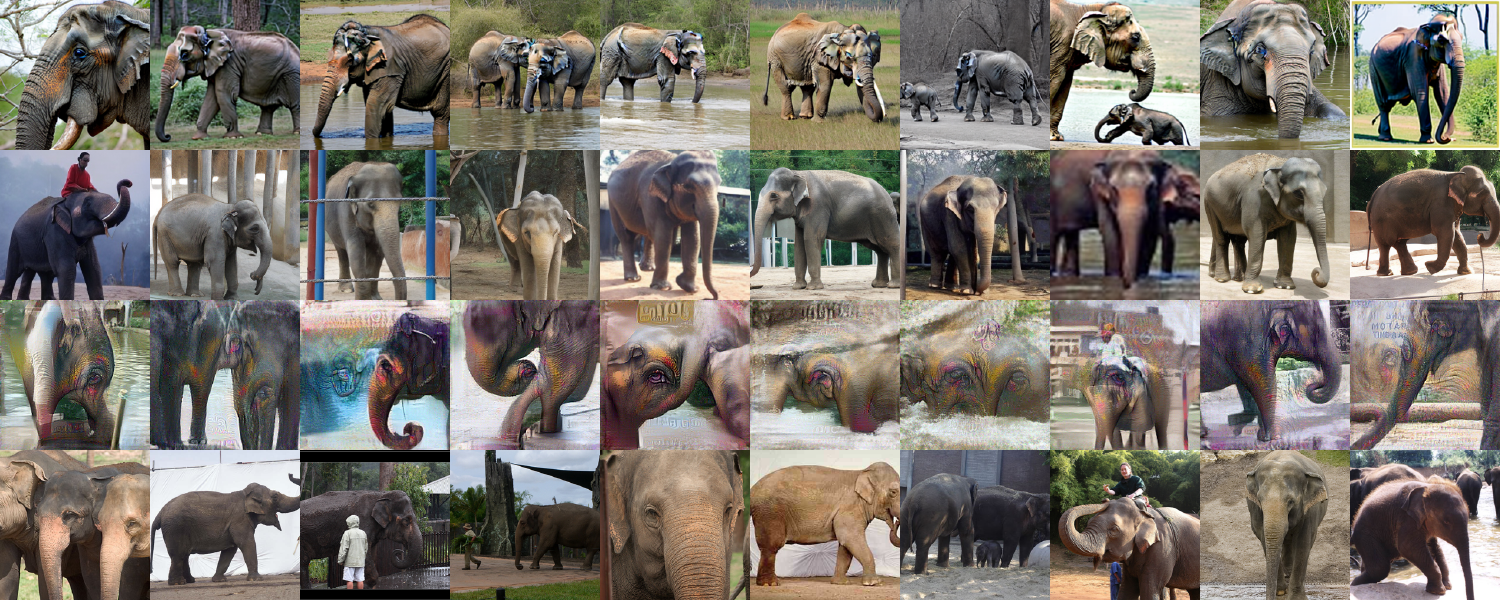}
        \caption{n02504013}
        \label{fig:elephant}
    \end{subfigure}
        \hfill
    \begin{subfigure}[t]{\linewidth}
        \centering
        \includegraphics[width=.85\linewidth]{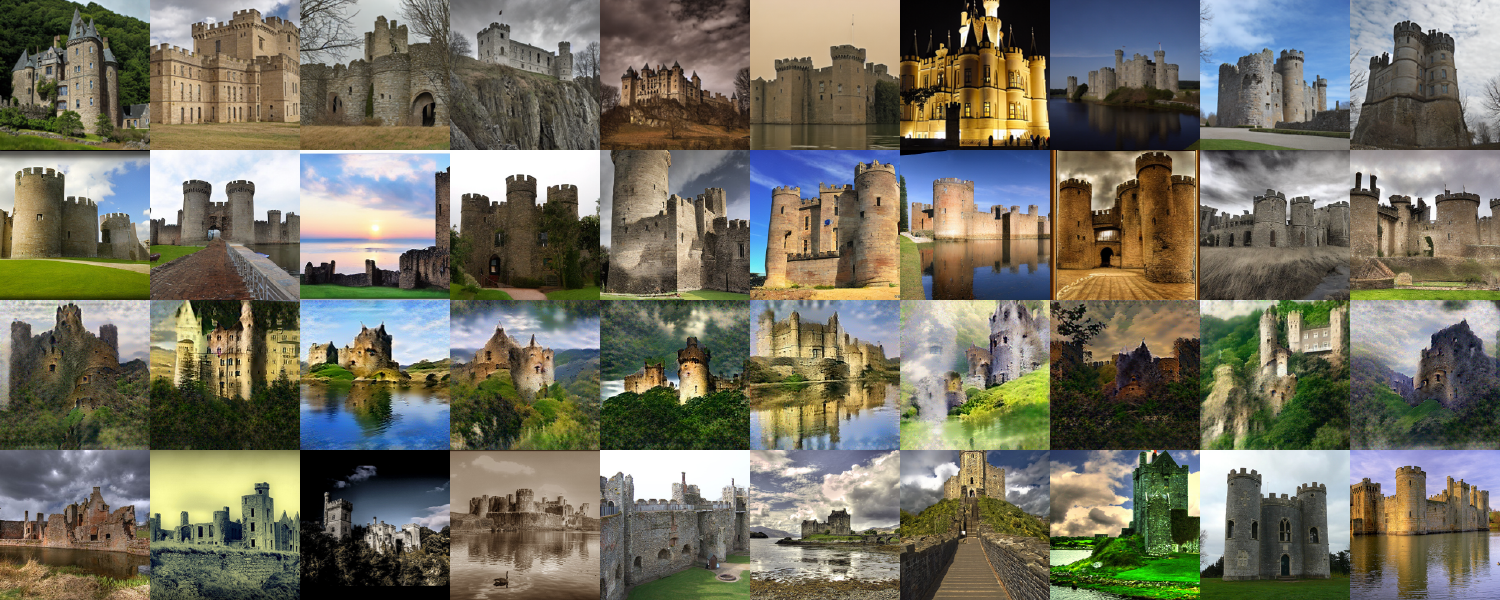}
        \caption{n02980441}
        \label{fig:castle}
    \end{subfigure}
    \caption{10 images per class sampled from condensed datasets produced by VLCP, ManifoldGD, FADRM+, and AUM for 4 synsets of ImageNet-1K.}
    \label{fig:grid_pg2}
\end{figure}

Figs.~\ref{fig:grid_pg1},~\ref{fig:grid_pg2} present grids showing ten samples per class generated or selected by four representative methods: VLCP, ManifoldGD, FADRM+, and AUM, across multiple ImageNet-1K classes. 
These visualizations complement the quantitative representativeness, diversity, and FID analyses by illustrating how different condensation strategies populate the condensed data space.

Across object‑centric classes such as lions and elephants, VLCP and ManifoldGD tend to repeatedly select or synthesize visually similar configurations, with limited variation in pose, background composition, or viewpoint. 
In many cases, identical object orientations or scene layouts appear multiple times within the same grid, indicating less coverage of intra-class variability.
While FADRM+ shows distinct poses or contexts, several images contain visible synthetic artifacts, including unnatural textures and distorted edges, particularly around object boundaries. 
This observation aligns with the reduced diversity and higher FID values reported for FADRM+.

In contrast, AUM consistently selects a broader range of real images, capturing variation in object scale, scene context, illumination, and background structure. 
For example, in fine‑grained bird and animal classes, AUM samples span multiple viewing angles and environmental settings, rather than repeatedly emphasizing a single canonical appearance. Similarly, for scene‑centric/structured object classes such as castles, AUM avoids near‑duplicate configurations observed in other methods. 
This visual coverage is consistent with AUM’s favorable position in the representativeness–diversity trade‑off and its lower FID scores.

Overall, these results highlight that DD condensation methods can often emphasize a narrow subset of visually salient modes, whereas CS methods such as AUM preserve a wider range of natural variations present in the original dataset.

\subsection{GitHub repositories referred}
\label{app:implementation_details}
In this work, we rely on the codes and distilled datasets provided by the following methods. We list their GitHub repositories here.
\begin{itemize}
    \item \textbf{DD methods}
    \begin{itemize}
        \item VLCP~\cite{zou2025vlcp}:
        \url{https://github.com/zou-yawen/Dataset-Distillation-via-Vision-Language-Category-Prototype}
        
        \item MGD$^3$~\cite{chan2025mgd3}:
        \url{https://github.com/jachansantiago/mode_guidance}
        
        \item ManifoldGD~\cite{roy2026manifoldgd}:
        \url{https://github.com/AyushRoy2001/ManifoldGD}
        
        \item D3HR~\cite{zhao2025taming}:
        \url{https://github.com/lin-zhao-resoLve/D3HR}
        
        \item FADRM+~\cite{cui2026fadrm}:
        \url{https://github.com/Jiacheng8/FADRM}
        
        \item Minimax~\cite{gu2024efficient}:
        \url{https://github.com/vimar-gu/MinimaxDiffusion}
        
        \item DiT Distillation~\cite{jin2024fast}:
        \url{https://github.com/chuanyangjin/fast-DiT}
    \end{itemize}

    \item \textbf{CS methods}
    \begin{itemize}
        \item CCS~\cite{zheng2022coverage}:
        \url{https://github.com/haizhongzheng/Coverage-centric-coreset-selection}
    \end{itemize}

    \item \textbf{Evaluation pipelines}
    \begin{itemize}
        \item RDED~\cite{sun2024diversityrealismdistilleddataset}:
        \url{https://github.com/LINs-lab/RDED}
        
        \item EDC~\cite{shao2024elucidating}:
        \url{https://github.com/shaoshitong/EDC}
    \end{itemize}
\end{itemize}

%%%%%%%%%%%%%%%%%%%%%%%%%%%%%%%%%%%%%%%%%%%%%%%%%%%%%%%%%%%%
\clearpage
\newpage
\section*{NeurIPS Paper Checklist}

\begin{enumerate}

\item {\bf Claims}
    \item[] Question: Do the main claims made in the abstract and introduction accurately reflect the paper's contributions and scope?
    \item[] Answer: \answerYes{} % Replace by \answerYes{}, \answerNo{}, or \answerNA{}.
    \item[] Justification: The abstract and introduction summarizes the paper’s main contributions.

\item {\bf Limitations}
    \item[] Question: Does the paper discuss the limitations of the work performed by the authors?
    \item[] Answer: \answerYes{} % Replace by \answerYes{}, \answerNo{}, or \answerNA{}.
    \item[] Justification: Limitations are discussed in Sec.~\ref{sec:conclusion}

\item {\bf Theory assumptions and proofs}
    \item[] Question: For each theoretical result, does the paper provide the full set of assumptions and a complete (and correct) proof?
    \item[] Answer: \answerNA{}. % Replace by \answerYes{}, \answerNo{}, or \answerNA{}.
    \item[] Justification: The paper does not include theoretical results.

    \item {\bf Experimental result reproducibility}
    \item[] Question: Does the paper fully disclose all the information needed to reproduce the main experimental results of the paper to the extent that it affects the main claims and/or conclusions of the paper (regardless of whether the code and data are provided or not)?
    \item[] Answer: \answerYes{} % Replace by \answerYes{}, \answerNo{}, or \answerNA{}.
    \item[] Justification: Details are in Secs.~\ref{sec:implementation_details},~\ref{sec:experiments} and the Appendix.

\item {\bf Open access to data and code}
    \item[] Question: Does the paper provide open access to the data and code, with sufficient instructions to faithfully reproduce the main experimental results, as described in supplemental material?
    \item[] Answer: \answerNo{} % Replace by \answerYes{}, \answerNo{}, or \answerNA{}.
    \item[] Justification: 
    Code release is pending approval and will be made public once approved. To support reproducibility, we fully specify the CS algorithms (AUM, Concepts, Clustering, IPC-CCS with the selected $\beta$ per dataset/IPC), and all DD distilled sets and evaluation pipelines (hard label, RDED, EDC) come from the official public repositories listed in App.~\ref{app:implementation_details}.
    
    % While the data and evaluation pipelines have been borrowed from open source GitHubs of prior works and appropriately cited in the main paper, we provide the evaluation scripts used for the three protocols in the our code submission. 

\item {\bf Experimental setting/details}
    \item[] Question: Does the paper specify all the training and test details (e.g., data splits, hyperparameters, how they were chosen, type of optimizer) necessary to understand the results?
    \item[] Answer: \answerYes{} % Replace by \answerYes{}, \answerNo{}, or \answerNA{}.
    \item[] Justification: Details are in Sec.~\ref{sec:experiments} and the Appendix.

\item {\bf Experiment statistical significance}
    \item[] Question: Does the paper report error bars suitably and correctly defined or other appropriate information about the statistical significance of the experiments?
    \item[] Answer: \answerYes{} % Replace by \answerYes{}, \answerNo{}, or \answerNA{}.
    \item[] Justification: All tables include standard deviation numbers for 3 runs. 

\item {\bf Experiments compute resources}
    \item[] Question: For each experiment, does the paper provide sufficient information on the computer resources (type of compute workers, memory, time of execution) needed to reproduce the experiments?
    \item[] Answer: \answerYes{} % Replace by \answerYes{}, \answerNo{}, or \answerNA{}.
    \item[] Justification: All the experiments were run on a machine with Intel(R) Xeon(R) Platinum797 with 16 CPUs with 64 GB of RAM and one Nvidia A-100 GPU.
    
\item {\bf Code of ethics}
    \item[] Question: Does the research conducted in the paper conform, in every respect, with the NeurIPS Code of Ethics \url{https://neurips.cc/public/EthicsGuidelines}?
    \item[] Answer: \answerYes{} % Replace by \answerYes{}, \answerNo{}, or \answerNA{}.
    \item[] Justification: Yes, our research conforms with the NeurIPS Code of Ethics.

\item {\bf Broader impacts}
    \item[] Question: Does the paper discuss both potential positive societal impacts and negative societal impacts of the work performed?
    \item[] Answer: \answerNA{} % Replace by \answerYes{}, \answerNo{}, or \answerNA{}.
    \item[] Justification: NA
    
\item {\bf Safeguards}
    \item[] Question: Does the paper describe safeguards that have been put in place for responsible release of data or models that have a high risk for misuse (e.g., pre-trained language models, image generators, or scraped datasets)?
    \item[] Answer: \answerNA{} % Replace by \answerYes{}, \answerNo{}, or \answerNA{}.
    \item[] Justification: NA

\item {\bf Licenses for existing assets}
    \item[] Question: Are the creators or original owners of assets (e.g., code, data, models), used in the paper, properly credited and are the license and terms of use explicitly mentioned and properly respected?
    \item[] Answer: \answerYes{} % Replace by \answerYes{}, \answerNo{}, or \answerNA{}.
    \item[] Justification: Our paper uses open-source data and models which have been appropriately cited.

\item {\bf New assets}
    \item[] Question: Are new assets introduced in the paper well documented and is the documentation provided alongside the assets?
    \item[] Answer: \answerNA{} % Replace by \answerYes{}, \answerNo{}, or \answerNA{}.
    \item[] Justification: We do not release any assets. % evaluation code is well documented.

\item {\bf Crowdsourcing and research with human subjects}
    \item[] Question: For crowdsourcing experiments and research with human subjects, does the paper include the full text of instructions given to participants and screenshots, if applicable, as well as details about compensation (if any)? 
    \item[] Answer: \answerNo{} % Replace by \answerYes{}, \answerNo{}, or \answerNA{}.
    \item[] Justification: We do not have such experiments.

\item {\bf Institutional review board (IRB) approvals or equivalent for research with human subjects}
    \item[] Question: Does the paper describe potential risks incurred by study participants, whether such risks were disclosed to the subjects, and whether Institutional Review Board (IRB) approvals (or an equivalent approval/review based on the requirements of your country or institution) were obtained?
    \item[] Answer: \answerNo{} % Replace by \answerYes{}, \answerNo{}, or \answerNA{}.
    \item[] Justification: We don’t have this type of studies.

\item {\bf Declaration of LLM usage}
    \item[] Question: Does the paper describe the usage of LLMs if it is an important, original, or non-standard component of the core methods in this research? Note that if the LLM is used only for writing, editing, or formatting purposes and does \emph{not} impact the core methodology, scientific rigor, or originality of the research, declaration is not required.
    %this research? 
    \item[] Answer: \answerNo{} % Replace by \answerYes{}, \answerNo{}, or \answerNA{}.
    \item[] Justification: LLMs are not a part of our pipeline.

\end{enumerate}

\end{document}